\documentclass[11pt]{article}
\usepackage{graphicx}

\usepackage[final]{acl}

\usepackage{times}
\usepackage{latexsym}

\usepackage[T1]{fontenc}

\usepackage[utf8]{inputenc}

\usepackage{microtype}

\usepackage{inconsolata}

\usepackage{graphicx}

\usepackage{amsmath,amsfonts,bm}

\def\eqref#1{equation~\ref{#1}}

\def\1{\bm{1}}

\DeclareMathAlphabet{\mathsfit}{\encodingdefault}{\sfdefault}{m}{sl}
\SetMathAlphabet{\mathsfit}{bold}{\encodingdefault}{\sfdefault}{bx}{n}

\usepackage{hyperref}
\usepackage{url}

\usepackage[utf8]{inputenc} 
\usepackage[T1]{fontenc}    
\usepackage{booktabs}       
\usepackage{amsfonts}       
\usepackage{nicefrac}       
\usepackage{microtype}      

\usepackage[T1]{fontenc}    
\usepackage{booktabs}       
\usepackage{amsfonts}       
\usepackage{nicefrac}       
\usepackage{microtype}      
\usepackage[table]{xcolor}
\definecolor{lightpink}{HTML}{FFF3F3}
\definecolor{lightblue}{HTML}{F3FAFF}
\definecolor{lightgray}{HTML}{F2F1F1}
\definecolor{darkgray}{HTML}{AEAEAE}
\usepackage{soul} 

\newcommand{\thinpinkcolorbox}[1]{{\fboxsep=1pt\fcolorbox{lightpink}{lightpink}{#1}}}
\newcommand{\thinbluecolorbox}[1]{{\fboxsep=1pt\fcolorbox{lightblue}{lightblue}{#1}}}
\newcommand{\thinlightgraycolorbox}[1]{{\fboxsep=1pt\fcolorbox{lightgray}{lightgray}{#1}}}
\newcommand{\thindarkgraycolorbox}[1]{{\fboxsep=1pt\fcolorbox{darkgray}{darkgray}{#1}}}

\usepackage{booktabs, multirow} 
\usepackage{soul}
\usepackage{graphicx}
\usepackage{wrapfig}
\usepackage{subcaption}
\usepackage{caption}
\usepackage{adjustbox}
\usepackage[most]{tcolorbox}
\usepackage{colortbl}
\usepackage{enumitem}
\newtcolorbox{prompt}[2][]{
    colback=white,
    colframe=gray!45,
    fonttitle=\bfseries,
    coltitle=black,
    sharp corners,
    title=#2,
    #1
}
\usepackage{longtable}
\newcommand{\aref}[1]{\hyperref[#1]{Appendix~\ref*{#1}}}

\usepackage[font=small]{caption}
\usepackage[bottom]{footmisc}

\usepackage{tcolorbox}
\tcbset{boxrule=0.1pt, arc=2mm, left=2mm, right=2mm, top=1mm, bottom=1mm}

\usepackage{color-edits}

\addauthor{gn}{magenta}

\usepackage{titletoc}

\title{Scaling Evaluation-Time Compute with Reasoning Models as Evaluators}

\author{Seungone Kim\textsuperscript{1}\thanks{denotes equal contribution.} \quad \textbf{Ian Wu}\textsuperscript{1*} \quad \textbf{Jinu Lee}\textsuperscript{2*} \quad \textbf{Xiang Yue}\textsuperscript{1} \quad \textbf{Seongyun Lee}\textsuperscript{3}\\
\textbf{Mingyeong Moon}\textsuperscript{4} \quad \textbf{Carolin Lawrence}\textsuperscript{5} \quad \textbf{Kiril Gashteovski}\textsuperscript{5,6} \quad \textbf{Julia Hockenmaier}\textsuperscript{2}\\ 
\textbf{Graham Neubig}\textsuperscript{1} \quad \textbf{Sean Welleck}\textsuperscript{1}\\
\\
\text{CMU}\textsuperscript{1} \quad \text{UIUC}\textsuperscript{2} \quad \text{KAIST AI}\textsuperscript{3} \quad \text{Independent Researcher}\textsuperscript{4}\\ 
\text{NEC Laboratories Europe}\textsuperscript{5} \quad \text{Ss.Cyril and Methodius University of Skopje}\textsuperscript{6} \quad\\
\\
\texttt{seungone@cmu.edu} \quad \texttt{ianwu@andrew.cmu.edu} \quad \texttt{jinulee2@illinois.edu}\\
}

\begin{document}

\maketitle

\begin{abstract}
Language model (LM) evaluators that generate chain-of-thought (CoT) reasoning are widely used for the assessment of LM responses.
Simultaneously, increasing LMs' ``thinking'' time through scaling test-time compute has proven to be an effective technique for solving challenging problems in domains such as math and code.
This raises a natural question: can an LM's evaluation capability also be improved by scaling test-time compute?
To answer this, we investigate employing \emph{reasoning models} -- LMs that natively generate long CoT reasoning -- as evaluators.
We explore scaling evaluation-time compute by using reasoning models to evaluate both the overall candidate response (\textit{i.e.}, outcome evaluation) and the individual reasoning steps within it (\textit{i.e.}, process evaluation). We observe that evaluator performance improves monotonically with the number of reasoning tokens generated, mirroring trends seen in LM reasoning. Furthermore, we use these more accurate evaluators to rerank multiple generations, and demonstrate that spending more compute at evaluation time can be as effective as increasing compute during generation for improving an LM's problem-solving performance.%
\end{abstract}

\section{Introduction}\label{sec:introduction}




Research on language models (LMs) involves an interplay between generation and evaluation: better generators require better evaluators and better evaluators can further enhance generators. For instance, an evaluator can verify the quality of the generator's response~\citep{liang2023holistic,zheng2024judging,ye2024flask} or identify the parts of a generator's response that contain mistakes~\citep{lightman2024let,zheng2024processbench,zhang2025lessons}. Furthermore, the generator's performance can be improved by integrating better evaluators into inference-time algorithms~\citep{cobbe2021training,uesato2022solving,lightman2024let,sun2024easytohard,wu2024better,zhang2024generative}.

\begin{figure}[!t]
    \centering
    \includegraphics[width=0.85\linewidth]{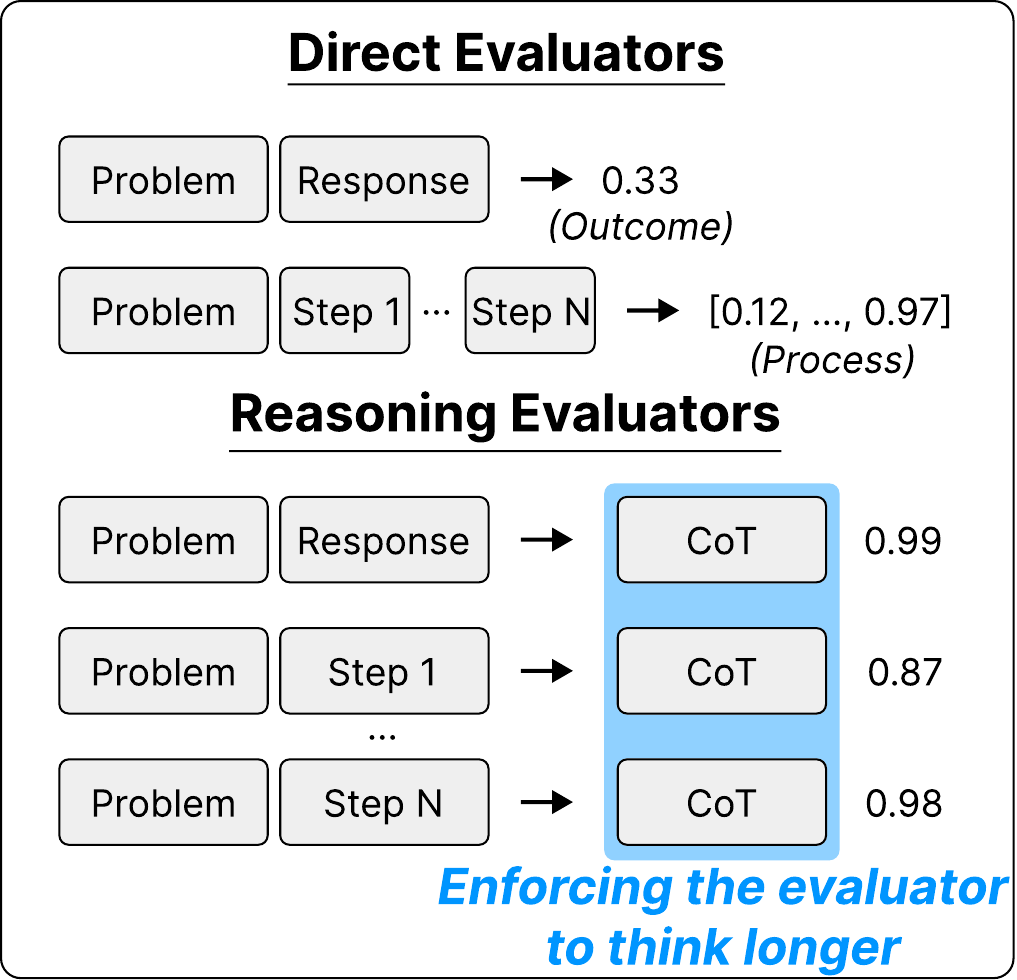}
    \caption{\textbf{We investigate the effect of scaling test-time compute for evaluation (evaluation-time compute)}: Enforcing the generation of additional reasoning tokens leads to improved evaluation performance (\autoref{section:scaling}). This can, in turn, be utilized to improve the generator’s performance (\autoref{section:problem_solving}).}
    \label{fig:taxonomy}
\end{figure}

Reasoning models have opened up a new paradigm for generation based on generating a long chain-of-thought (CoT)~\citep{jaech2024openai,guo2025deepseek,muennighoff2025s1,aggarwal2025l1}. Prior works have found that generating long CoTs is an effective strategy for leveraging test-time compute to solve difficult tasks that conventional instruction-tuned models cannot ~\citep{yeo2025demystifying,xu2025towards}. 
However, it is unclear whether evaluators, like generators, can also be improved by scaling test-time compute with long CoTs.
In this paper, we ask and answer two questions: (1) Can we replicate test-time scaling behavior observed in generators with evaluators? And if so, (2) can this improved evaluation ability further improve generation results as well?
Our main contribution is an examination of generative evaluators that use long CoT reasoning, which we refer to as \textit{reasoning evaluators}. Reasoning evaluators are obtained by prompting reasoning models -- a class of models trained to produce long CoT either using RL or through distillation from the outputs of such a model~\citep{guo2025deepseek, qwq-32b-preview} -- to act as evaluators. By generating long outputs exhibiting complex reasoning patterns including self-verification and backtracking~\citep{gandhi2025cognitivebehaviorsenableselfimproving, lu2025retrosearchexploringuntakenpaths}, reasoning models expend test-time compute to attain improved reasoning capabilities, a trait that we hypothesize also makes them better evaluators. In this work, we contrast reasoning evaluators with (1) \textit{direct evaluators}, which predict scores without CoT reasoning~\citep{lightman2024let, cobbe2021training}, and (2) \textit{fine-tuned generative evaluators} that produce shorter CoT lacking the complex reasoning patterns found in reasoning model outputs~\citep{zhang2024generative, ankner2024critique, kim2024prometheus}.  

Specifically, we force reasoning evaluators to generate more reasoning tokens by prompting them to evaluate both each step of an output individually as well as the solution as a whole.
This recipe unifies techniques from prior work on step-by-step evaluation (process reward models; PRMs; \citep{lightman2024let,wang2024math}) and outcome-based evaluation (outcome reward models; ORMs; \citep{cobbe2021training,liu2024skywork}).

We demonstrate the effectiveness of our approach across two settings. First, we show that the evaluator's performance improves monotonically as it generates more reasoning tokens. We further show that a 32B reasoning evaluator can outperform a 72B state-of-the-art PRM by a 4.5\% margin on ProcessBench~\citep{zheng2024processbench}, a benchmark that measures whether an LM can identify the first occurring error within a given response. This is notable because, while existing direct evaluators are trained on extensive process supervision, reasoning evaluators achieve strong performance without any training, relying solely on test-time scaling (referred to as `evaluation-time scaling'). 

Second, we find that evaluation-time scaling is an effective method for further improving the generator's performance. When integrating reasoning evaluators into Best-of-$N$ sampling, where an evaluator reranks multiple solutions sampled by a generator, our reasoning evaluators using Best-of-8 outperform direct evaluators (\textit{e.g.}, ORMs, PRMs) using Best-of-64 by a 4.30 - 6.63\% margin given a fixed compute budget, highlighting the benefits of spending more test-time compute for evaluation at the expense of sampling more responses.

\section{Methodology}\label{sec:gen_process_evaluator}


We describe our approach for scaling evaluation-time compute by assessing both overall responses (\autoref{subsec:orm}) and individual response segments (\autoref{subsec:prm}) with reasoning evaluators. We then explain how we combine process and outcome judgments (\autoref{subsec:beam}) for further gains.

\textbf{Reasoning Evaluators vs. Direct Evaluators} We refer to conventional evaluators that are trained to map a problem and a response (or steps) to a scalar value score as \emph{direct evaluators}. Reasoning evaluators differ from direct evaluators in two aspects. First, reasoning evaluators generate chain-of-thought (CoT) reasoning before predicting the final judgment. Second, while direct evaluators with a specialized reward modeling head must be fine-tuned, reasoning evaluators may either be specifically trained for evaluation\footnote{We also classify trained evaluators such as CLoud~\citep{ankner2024critique} and Prometheus~\citep{kim2024prometheus2} as reasoning evaluators, although they produce relatively short CoTs that lack the versatile reasoning patterns seen in reasoning models outputs.} or may be off-the-shelf LMs that are prompted to act as evaluators. In this paper, we focus on the latter approach by prompting reasoning models to function as evaluators.

Given a problem $x_i$ and response $y_i$, the evaluator is used to estimate the ``goodness'' of $y_i$ by generating a score $s_i$. This score can be obtained with a trained reward modeling head (direct evaluators) or the logits of answer tokens (\textit{e.g.}, 0/1) (reasoning evaluators). The mapping function of outcome and process evaluators are expressed as

\begin{itemize}[leftmargin=*]
    \item \textbf{Outcome Evaluator}: $(x_i, y_i) \rightarrow s_i$
    \item \textbf{Process Evaluator}:\\ $(x_i, [y_{i1}, y_{i2}, ..., y_{iN}]) \rightarrow [s_{i1}, s_{i2}, ..., s_{iN}]$.
\end{itemize}
Process evaluators require a \textbf{splitting function} to divide $y_i$ into discrete steps $[y_{i1}, y_{i2}, ..., y_{iN}]$. Furthermore, process evaluators can only be used for evaluation of the final outcome if provided with an \textbf{aggregation function} that maps per-step scores $[s_{i1}, s_{i2}, ..., s_{iN}]$ to an aggregated final score $s_i$. Conventionally, a heuristic-based approach is used as the splitting function (\textit{e.g.}, splitting based on ``\texttt{$\backslash$n$\backslash$n}'') while the \texttt{min} function ($s_i = min(s_{i1}, s_{i2}, ..., s_{iN})$) is often used as the aggregation function~\citep{lightman2024let,wang2024math,sun2024easytohard}. As previously discussed, direct outcome and process evaluators predict these values through specially trained heads. 

\begin{figure*}[t]
    \centering
    \includegraphics[width=0.9\linewidth]{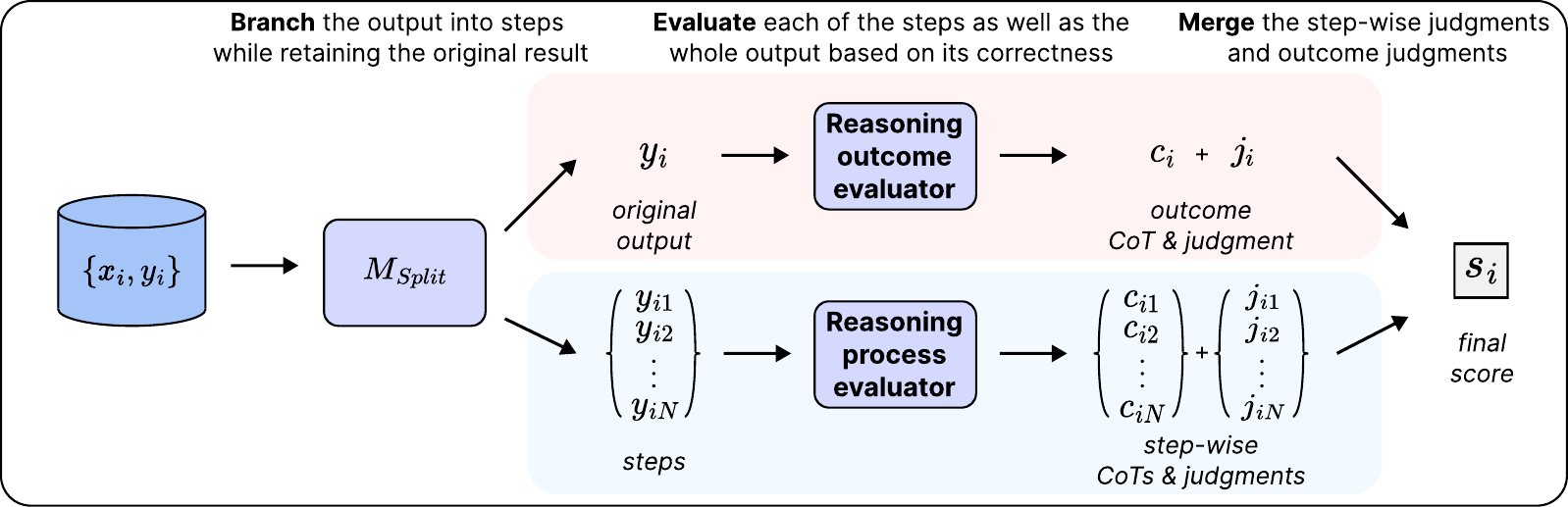}
    \caption{We propose scaling evaluation-time compute by using the evaluator to assess both the overall output (light red) and its constituent reasoning steps (light blue) and then combining the judgments into a final score.}
    \label{fig:beam}
\end{figure*}

In the following subsections we discuss how to make these predictions with reasoning models.

\subsection{Reasoning Outcome Evaluators}\label{subsec:orm}

Our reasoning outcome evaluators have at their core a function
\begin{equation}
    (x_i, y_i) \rightarrow (c_i, j_i),
\end{equation}
where $c_i$ denotes a CoT, and $j_i$ is the evaluator's judgment, represented as the probability distribution over tokens in the vocabulary: see the upper section of \autoref{fig:beam} (colored light red).
We prompt the LM to output ``1'' if the response is deemed to be correct and ``0'' if not. To transform $j_i$ into a scalar value score $s_i$, we use the logits $\ell$ of ``1'' and ``0'' tokens and perform a softmax operation as
\begin{equation}\label{equation:softmax}
    s_{i} = \frac{e^{\ell(j_{i}=1)}}{e^{\ell(j_{i}=0)}+e^{\ell(j_{i}=1)}}.
\end{equation}


\subsection{Reasoning Process Evaluators}\label{subsec:prm}

We formulate the mapping function for assessing reasoning step $k$ as
\begin{equation}
    (x_i, [y_{i1}, ..., y_{ik}]) \rightarrow (c_{ik}, j_{ik})\; \; \; \; (1 \leq k \leq N),
\end{equation}
where $c_{ik}$ denotes the CoT that examines $y_{ik}$ for potential logical flaws or inconsistencies and $j_i$ denotes the judgment for $y_{ik}$, which is also represented as a probability distribution: see the bottom section of \autoref{fig:beam} (colored light blue). Note that the previous steps $[y_{i1}, ..., y_{i(k-1)}]$ are provided as context for precise assessment of the current step $y_{ik}$.
Then, to convert $j_{ik}$ into $s_{ik}$, we use

\begin{equation}\label{equation:softmax2}
\begin{split}
    [s_{i1}, ..., s_{iN}] = \Bigg[&\frac{e^{\ell(j_{i1}=1)}}{e^{\ell(j_{i1}=0)}+e^{\ell(j_{i1}=1)}}, ..., \\
    &\frac{e^{\ell(j_{iN}=1)}}{e^{\ell(j_{iN}=0)}+e^{\ell(j_{iN}=1)}}\Bigg].
\end{split}
\end{equation}

\textbf{Single- vs. multi-step process evaluation} The formulation above forms the core of our proposed method, but we also compare with an ablated \emph{single-step} reasoning process evaluator proposed by \citet{zheng2024processbench} that generates a single CoT before making judgments of all $N$ steps in $y_i$ as
\begin{equation}
    (x_i, [y_{i1}, ..., y_{iN}]) \rightarrow (c_{i}, [j_{i1}, \ldots, j_{iN}]).
\end{equation}

Unless explicitly stated, reasoning process evaluation refers to our proposed multi-step variant.
Evaluating each step \textit{separately} is our preferred method for two reasons: (1) evaluating all $N$ steps at once risks exceeding the context window of reasoning models, and (2) stepwise evaluation forces the evaluator to assess each step more thoroughly, thereby naturally scaling evaluation-time compute.

\textbf{Choice of splitting function and aggregation function} Additionally, we make the following adjustments to the splitting and aggregation functions when using reasoning process evaluators:

\begin{itemize}[leftmargin=*]
    \item \textbf{Model-based splitting}: When splitting $y_i$ into $[y_{i1}, ..., y_{iN}]$, conventional heuristic-based approaches may be ineffective for some cases (\textit{e.g.}, when $y_i$ does not include ``\texttt{$\backslash$n$\backslash$n}'' or is not written in a structured format, as is the case for code). To deal with this, we adopt model-based splitting where an LM $M_{split}$ is prompted to insert an indicator phrase ``\texttt{[SPLIT]}'' between steps as
\begin{equation}
    M_{split}: y_i \rightarrow [ y_{i1}\; \texttt{[SPLIT]}\; ...\; \texttt{[SPLIT]}\; y_{iN} ].
\end{equation}

\item \textbf{Score aggregation}: After acquiring $[s_{i1},  ..., s_{iN}]$ as in \autoref{equation:softmax2}, we aggregate the $N$ judgments into a single scalar value score $s_i$. In our experiments, the \texttt{mean$\_$logit} function~\citep{sun2024easytohard} yields better results than \texttt{min} function. The \texttt{mean$\_$logit} function is expressed as

\begin{equation}
\begin{split}
    s_i &= \texttt{mean\_logit}(s_{ik}) \\
        &= \sigma\left(\frac{\sum_{k}\frac{s_{ik}}{1 - s_{ik}}}{N}\right) \quad (1 \leq k \leq N).
\end{split}
\end{equation}

\end{itemize}

Note that these adjustments can be applied to direct process evaluators as well: see \autoref{appendix:bon_extended}.



\subsection{Combining outcome judgments and process scores}\label{subsec:beam}

While the objective of outcome evaluation is to determine the \textit{correctness of the final answer}, the objective of process evaluation is to determine the \textit{correctness of each step}. 
Both have their advantages in identifying reasoning errors -- outcome evaluation takes a more holistic approach while process evaluation can potentially identify more fine-grained errors. Inspired by \citet{uesato2022solving}, we consider combining both outcome and process evaluation scores through interpolation as follows
\begin{equation}
    \label{eq:beam}
    s_{final} = \alpha \cdot s_{outcome} + (1-\alpha) \cdot s_{process}.
\end{equation}


Here, choosing $\alpha=0$ is identical to only using the process score and choosing $\alpha=1$ is identical to only using the outcome score. We use $\alpha = 0.5$ to avoid overfitting to either approach and refer this method as \textbf{reasoning process + outcome evaluation}. See \autoref{appendix:why} for effect on varying $\alpha$.


\begin{table*}[!t]
\tiny
\centering
\resizebox{0.9\textwidth}{!}{
\begin{tabular}{lcccccc}
\toprule
\textbf{Model} & \textbf{GSM8K} & \textbf{MATH} & \textbf{Olym. Bench} & \textbf{Omni-MATH} & \textbf{Avg. F1} & \textbf{$\Delta$}\\
\midrule
\multicolumn{7}{c}{\textbf{Direct Process Evaluator (PRMs)}} \\
\midrule
Qwen2.5-Math-7B-PRM800K & 68.4 & 62.5 & 50.4 & 43.6 & 56.2 & - \\
Qwen2.5-Math-PRM-7B &   82.4 	& 77.6		& 67.5	& 66.3	& 73.5 & -\\
Qwen2.5-Math-PRM-72B &  \textbf{87.3}	& \textbf{80.6}& \textbf{74.3}		& \textbf{71.1}	& \textbf{78.3} & -\\
\midrule
\multicolumn{7}{c}{\textbf{Single-step Reasoning Process Evaluator}} \\
\midrule
\noalign{\vspace{2pt}}
\multicolumn{7}{l}{\textbf{Instruction-tuned Models}} \\
Llama-3.1-8B-Instruct   & 24.0& 15.5 & 9.7  & 10.1 & 14.8 & 0.0\\
Qwen2.5-32B-Instruct & 63.8 & 47.5 & 35.9 & 32.7 & 45.0 & 0.0\\
Qwen2.5-72B-Instruct  & 76.2 & 61.8& 54.6 & 52.2 & 61.2 & -\\
\noalign{\vspace{2pt}}
\multicolumn{7}{l}{\textbf{Reasoning Models}} \\
\rowcolor{lightgray} DeepSeek-R1-Distill-Qwen-7B  & 68.3 & 61.1 & 48.3 & 40.4 &54.5 & 0.0\\
\rowcolor{lightgray} DeepSeek-R1-Distill-Qwen-32B  & 83.9 & 78.1 & 72.4 & 67.7 & 75.5 & 0.0\\
\rowcolor{lightgray} QwQ-32B-Preview   & 77.5 & 58.9 & 31.2  & 35.8 & 50.9 & 0.0\\
\rowcolor{lightgray} QwQ-32B   & 79.5 & 77.5 & 71.5  & 69.4 & 74.5 & 0.0\\
\noalign{\vspace{2pt}}
\multicolumn{7}{l}{\textbf{Reasoning Models} (Self-Consistency)} \\
\rowcolor{lightgray} DeepSeek-R1-Distill-Qwen-7B & 69.3&67.9&54.8 & 51.5 & 60.9 & +6.4\\
\rowcolor{lightgray} DeepSeek-R1-Distill-Qwen-32B & 82.2 & \textbf{80.4} & \textbf{76.2} & 72.5 & \textbf{77.8} & +2.3\\
\rowcolor{lightgray} QwQ-32B-Preview & \textbf{88.0} & 78.7 & 57.8& 61.3 & 71.5 & +20.6\\
\rowcolor{lightgray} QwQ-32B & 81.0 & 78.8 & 74.4 & \textbf{72.8} & 76.8 & +2.3\\
\noalign{\vspace{2pt}}
\midrule
\multicolumn{7}{c}{\textbf{Reasoning Process Evaluator (Ours)}} \\
\midrule
\noalign{\vspace{2pt}}
\multicolumn{7}{l}{\textbf{Instruction-tuned Models} (Multi-step Process Evaluation)} \\
Llama-3.1-8B-Instruct & 35.2& 22.8  & 12.6  & 17.9 & 22.1&+7.3\\
Qwen2.5-32B-Instruct& 70.1 & 61.7 & 54.2 & 53.9 & 60.0 & +15.0\\
\noalign{\vspace{2pt}}
\multicolumn{7}{l}{\textbf{Reasoning Models} (Multi-step Process Evaluation)} \\
\rowcolor{lightblue} DeepSeek-R1-Distill-Qwen-7B & 75.5 & 67.3 & 59.8 & 56.6 & 64.8&+10.3\\
\rowcolor{lightblue} DeepSeek-R1-Distill-Qwen-32B &  80.3 & 82.2 & 77.0 & 75.0 & 78.6 & +3.1\\
\rowcolor{lightblue} QwQ-32B-Preview  & 81.7 & 79.3 & 70.3 & 69.8 & 75.3 & +24.4\\
\rowcolor{lightblue} QwQ-32B  & 81.5 & 83.6 & 76.8 & 75.1 & 79.3 & +4.8\\
\noalign{\vspace{2pt}}
\multicolumn{7}{l}{\textbf{Reasoning Models} (Multi-step Process Evaluation + Self-Consistency)} \\
\rowcolor{lightblue} DeepSeek-R1-Distill-Qwen-7B & 80.1& 75.1 & 69.3 & 70.4 & 73.7 & +19.2\\
\rowcolor{lightblue} DeepSeek-R1-Distill-Qwen-32B & 86.6 & 85.4 & 78.9 & \textbf{80.3} & \textbf{82.8} & +7.3\\
\rowcolor{lightblue} QwQ-32B-Preview &  \textbf{86.8} & \textbf{85.7}  & 79.0 & 78.0 & 82.4&+31.5\\
\rowcolor{lightblue} QwQ-32B &  85.0 & 85.6 & \textbf{79.4} & 78.0 & 82.0 & +7.5\\
\bottomrule
\end{tabular}
}
\caption{\textbf{Experimental results for ProcessBench experiments}. Scaling evaluation-time compute through multi-step process evaluation and self-consistency improves performance, with even non-fine-tuned LMs outperforming specially trained PRMs. $\Delta$ denotes gains associated with either applying (1) multi-step process evaluation or (2) self-consistency for the same backbone LM. The best performance within each category is \textbf{bolded}. See \autoref{tab:processbench_extended} for additional results.}
\label{tab:processbench}
\vspace{-3mm}
\end{table*}

\section{Evaluation-Time Scaling Trends for Process Evaluation}\label{section:scaling}

As discussed in \autoref{sec:gen_process_evaluator}, our approach involves assessing output steps \textit{individually} (multi-step process evaluation) using reasoning evaluators. To examine the effectiveness of this choice, we experiment on a response error detection task and compare our method against state-of-the-art PRMs. 

\subsection{Experimental setting}\label{subsec:processbench}

\textbf{Benchmark} We explore scaling evaluation-time compute with ProcessBench~\citep{zheng2024processbench}, which includes diverse responses from different LMs and highly reliable human-annotated labels. In this setting. evaluators are tasked with identifying the first paragraph in the solution that contains incorrect logic, if any. ProcessBench has 3,400 instances, with queries sampled from 4 different math benchmarks and responses from 12 distinct LMs. Each response consists of 7.56 steps on average.

\textbf{Metric} The evaluator's performance on ProcessBench is measured using the F1 score, computed from the precision and recall of predicting the index of the first paragraph that contains a logical error: evaluators are penalized for incorrectly identifying a paragraph as erroneous when no error exists, misidentifying the index of the erroneous paragraph, or failing to detect an error when one is present. If the evaluator predicts that one or more steps are incorrect, we use the earliest incorrect step (the step with the smallest index) as the final prediction, following \citet{zheng2024processbench}.

\textbf{Methods} We consider the following evaluator baselines (see Appendix~\ref{appendix:model_list_scaling} for the list of models):

\begin{itemize}[leftmargin=*]
    \item \textbf{Direct process evaluator}: We employ process reward models (PRMs) as direct process evaluators. These models \textit{do not} generate CoTs but instead directly predict the correctness of all steps. 
    \item \textbf{Single-step reasoning process evaluator}: We adopt the approach proposed by \citet{zheng2024processbench} and \citet{zhang2025lessons}, where a language model is given the response and prompted to produce a single CoT as well as the index of the first paragraph containing an error, if one exists. This corresponds to the ablated ``single-step'' evaluator discussed in \autoref{subsec:prm}.
    \item \textbf{Reasoning process evaluator}: We explore our approach for using reasoning models as process evaluators. This involves assessing each segment \textit{individually} and determining its correctness.
    For this experiment, we use a simplified version of the approach described in \autoref{subsec:prm}:  (1) a splitting function is not required as segments are already provided and (2) an aggregation function is not required as the goal of the benchmark is to predict the index of the segment containing the first mistake, not to produce an outcome score to compare multiple responses.
\end{itemize}


\textbf{Matching test-time budget across methods} To ensure a fair comparison between single-step evaluators and our method under similar test-time compute constraints, we also experiment with \textbf{self-consistency}~\citep{wang2023selfconsistency} for evaluation. Specifically, the evaluator generates $M$ CoT trajectories (\textit{e.g.}, if applied to our reasoning process evaluator, it assesses each of $N$ steps $M$ times, resulting in a total of $N \cdot M$ inference calls), with $j_{ik}$ chosen based on a majority vote. Self-consistency is inapplicable for PRMs since they produce identical scores across multiple inference calls.

\subsection{Experimental results}

Our main results on ProcessBench are presented in \autoref{tab:processbench}. See \autoref{tab:processbench_extended} for additional results.

\textbf{Finding 1: Reasoning models are better evaluators than instruction-tuned models}. For example, DeepSeek-R1-Distill-Qwen-32B achieves an average F1 score of 75.5 when employed as a reasoning process evaluator, significantly outperforming the larger Qwen2.5-72B-Instruct model (61.2 F1) despite having only 44\% as many parameters. This suggests that reasoning capability is associated with improved evaluation capability, even when the reasoning model is not trained as an evaluator.

\textbf{Finding 2: Single-step methods fail to match the performance of direct process evaluators}. For example, DeepSeek-R1-Distill-Qwen-7B achieves an F1 score of 54.5, which is lower than Qwen2.5-Math-7B-PRM800K (56.2) and Qwen2.5-Math-PRM-7B (73.5). Similarly, DeepSeek-R1-Distill-Qwen-32B (75.5) and QwQ-32B (74.5) fall behind the Qwen2.5-Math-PRM-72B (78.3). This indicates that single-step evaluation is not sufficient to identify errors in responses.

\textbf{Finding 3: It is important to evaluate each step individually (multi-step process evaluation)}. As detailed in the previous subsection, we compare two approaches for scaling evaluation-time compute: (1) self-consistency, where the evaluator generates multiple CoTs for evaluation and the final judgment is decided via majority vote and (2) multi-step process evaluation, where the evaluator assesses each step in the solution individually. As ProcessBench has an average of 7.56 steps and we generate 8 CoTs for the self-consistency baseline, the two approaches incur similar inference costs. 

Our results indicate that prompting reasoning models to evaluate each step individually is more effective than applying self-consistency given a fixed evaluation compute budget. Using multi-step process evaluation, DeepSeek-R1-Distill-Qwen-{7,32}B, QwQ-32B-Preview, and QwQ-32B achieve F1 scores of 64.8, 78.6, 75.3, and 79.3, respectively. In contrast, applying self-consistency yields lower scores of 60.9, 77.8, 71.5, and 76.8. Notably, our multi-step process evaluation method enables DeepSeek-R1-Distill-Qwen-32B (78.6) and QwQ-32B (79.3) to outperform Qwen2.5-Math-PRM-72B (78.3) (a model nearly twice as big) without any additional training. Furthermore, instruction-tuned models also achieve sizable gains when used for multi-step process evaluation, with Llama-3.1-8B-Instruct and Qwen2.5-32B-Instruct demonstrating gains of +7.3 and +15.0 respectively. These results indicate that our method can be applied to non-reasoning LMs as well.


\textbf{Finding 4: Combining self-consistency and multi-step process evaluation can further enhance performance}. We also find that further scaling evaluation-time compute by applying multi-step process evaluation and self-consistency together yields even more gains. For instance, DeepSeek-R1-Distill-Qwen-7B (73.7) outperforms Qwen2.5-Math-PRM-7B (73.5). Similarly, DeepSeek-R1-Distill-Qwen-32B (82.8) and QwQ-32B (82.0) surpass Qwen2.5-Math-PRM-72B (78.3), which was the previous state-of-the-art~\citep{zhang2025lessons}, suggesting that both self-consistency and multi-step process evaluation can provide complementary benefits for scaling evaluation-time compute. This is notable because the reasoning models are not explicitly trained as evaluators, yet increasing evaluation-time compute consistently improves performance and achieves state-of-the-art results.




\begin{figure*}[t]
    \centering
    \includegraphics[width=0.85\linewidth]{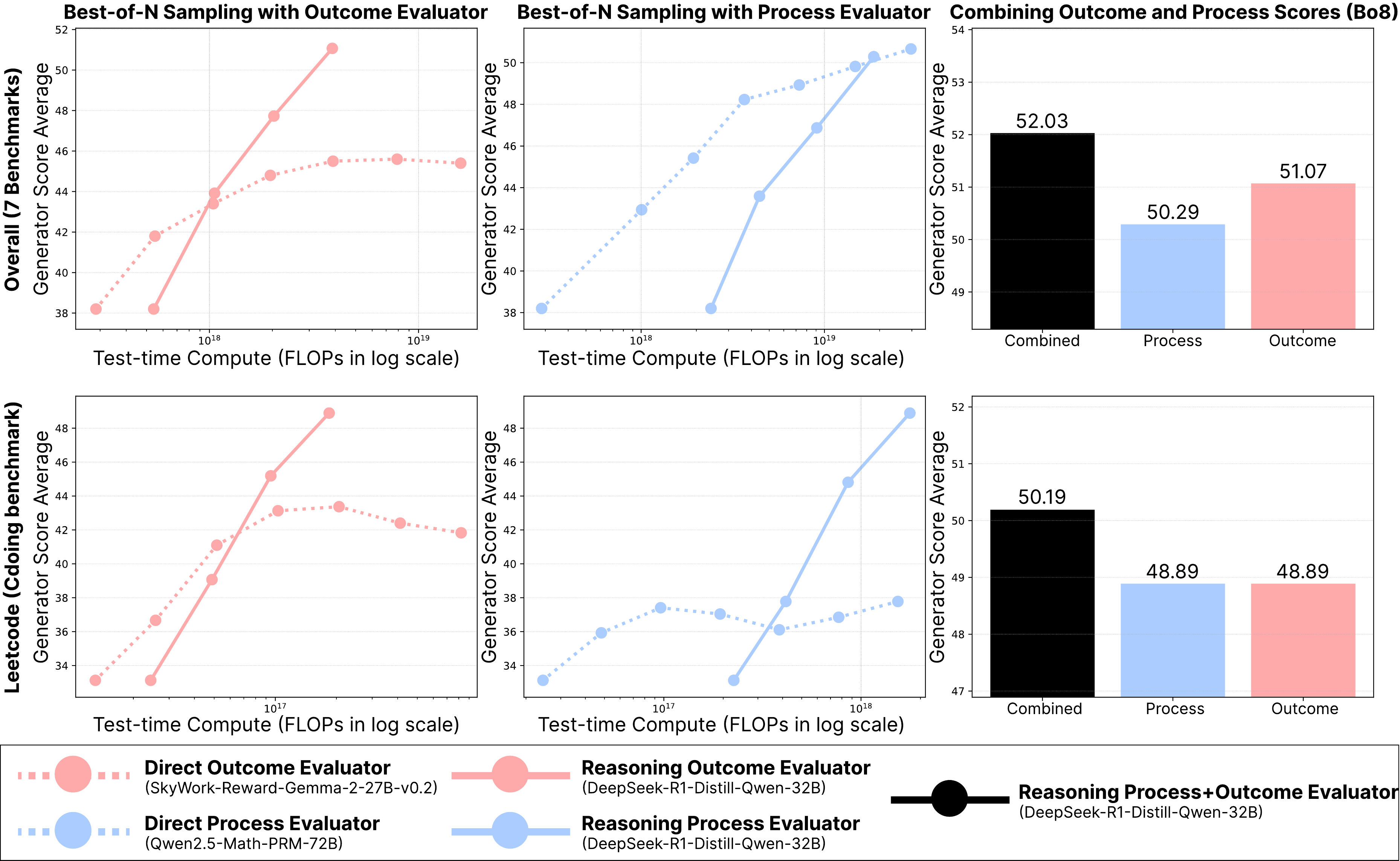}
    \caption{\textbf{Best-of-$N$ experimental results for 27B $\sim$ 72B scale evaluators}. Test-time compute on the x axis denotes the generator's test-time compute added with evaluation-time compute. \textbf{(left, middle)} We compare direct evaluators (using Best-of-64) against reasoning evaluators (using Best-of-8), with each dot representing a doubling of the number of responses. We find that reasoning evaluators outperform their direct counterparts given a fixed test-time compute budget: see \autoref{sec:appendix-approx-compute} for details of our FLOPs calculations and \autoref{tab:bon_extended1} for full results. \textbf{(right)} Combining outcome and process scores yields further gains.
    }
    \label{fig:bon1}
    \vspace{-3mm}
\end{figure*}

\section{Translating Improved Evaluation to Problem-Solving}\label{section:problem_solving}


Building upon the previous section, we explore whether we can leverage evaluation-time scaling to improve the generator's performance. We investigate this with Best-of-$N$ sampling, where test-time scaling is conventionally achieved by generating more response candidates. In contrast, we explore whether allocating more evaluation-time compute at the expense of sampling fewer candidate responses can ultimately yield better performance given a fixed test-time compute budget.

\subsection{Experimental setting}\label{subsec:best_of_n}

\textbf{Benchmarks} We follow the setting of \citet{cui2025process} by utilizing three LMs as generators, namely Eurus-2-SFT~\citep{cui2025process}, Llama3.1-70B-Instruct~\citep{dubey2024llama3herdmodels}, and Qwen2.5-7B-Instruct~\citep{yang2024qwen2}. We generate 64 responses per instance across seven benchmarks (AIME24, AMC23, Minerva Math~\citep{lewkowycz2022solving}, OlympiadBench~\citep{he2024olympiadbench}, MATH500~\citep{hendrycks2021measuring}, LeetCode~\citep{guo2024deepseek}, and GPQA~\citep{rein2024gpqa}), resulting in 4,680 instances and 299,520 responses. In Best-of-$N$ setting, evaluators assess and rank the $N$ responses, with the highest-scoring response used as the final prediction.

\textbf{Metrics} For LeetCode, we report pass@1, which measures whether a response passes all test cases. For the remaining 6 benchmarks, we report accuracy scores, which measures whether a response is correct. We report the average score of the 3 generators across all 7 benchmarks (21 settings). 

\textbf{Methods} We consider the following evaluator baselines (see Appendix~\ref{appendix:model_list_bon} for the list of models):


\begin{itemize}[leftmargin=*]
    \item \textbf{Direct outcome evaluator}: We employ outcome reward models (ORMs) as direct outcome evaluators. Note that these models do not generate CoTs but instead directly predict the scores.
    \item \textbf{Direct process evaluator}: We employ process reward models (PRMs) as direct process evaluators. We adopt the splitting functions and aggregation strategies accompanying each PRM’s official implementation, as specified in their GitHub repositories or Hugging Face model cards. Note that these models do not generate CoTs but instead directly predict the correctness of all steps.
    \item \textbf{Reasoning outcome evaluator}: We prompt reasoning models to act as reasoning outcome evaluators. The evaluator first generates a CoT, followed by a judgment for the correctness of the overall response. Details are in \autoref{subsec:orm}.
    \item \textbf{Reasoning process evaluator}: We prompt reasoning models to act as reasoning process evaluators. The evaluator individually generates a CoT as a judgment for each step. We use our proposed model-based splitting strategy and the \texttt{mean\_logits} function (see \autoref{subsec:prm}). 
    \item \textbf{Reasoning process + outcome evaluator}: Using the judgments from a reasoning outcome evaluator and a reasoning process evaluator, we combine scores to obtain the overall score for the response. Details are included in \autoref{subsec:beam}. We use $\alpha = 0.5$ to avoid overfitting. 
\end{itemize}

In \autoref{appendix:model_list_bon}, we experiment with generative evaluators that were fine-tuned to function as outcome evaluators, and also study the effects of applying model-based splitting and \texttt{mean\_logits} aggregation to direct process evaluators. Prompt templates can be found in \autoref{appendix:reasoning_evaluator_prompts}.

\textbf{Matching evaluation-time budget across baselines} We test direct outcome evaluators and direct process evaluators in the Best-of-64 setting. To account for the higher per-instance inference cost of reasoning evaluators, we evaluate them in the Best-of-8 setting instead, thereby ensuring a compute budget comparable to that of direct evaluators. We use \texttt{Qwen2.5-72B-Instruct} to segment the response into steps, yielding 10.07 steps per response on average across the 21 settings.

\begin{figure*}[!t]
    \centering
    \includegraphics[width=0.8\linewidth]{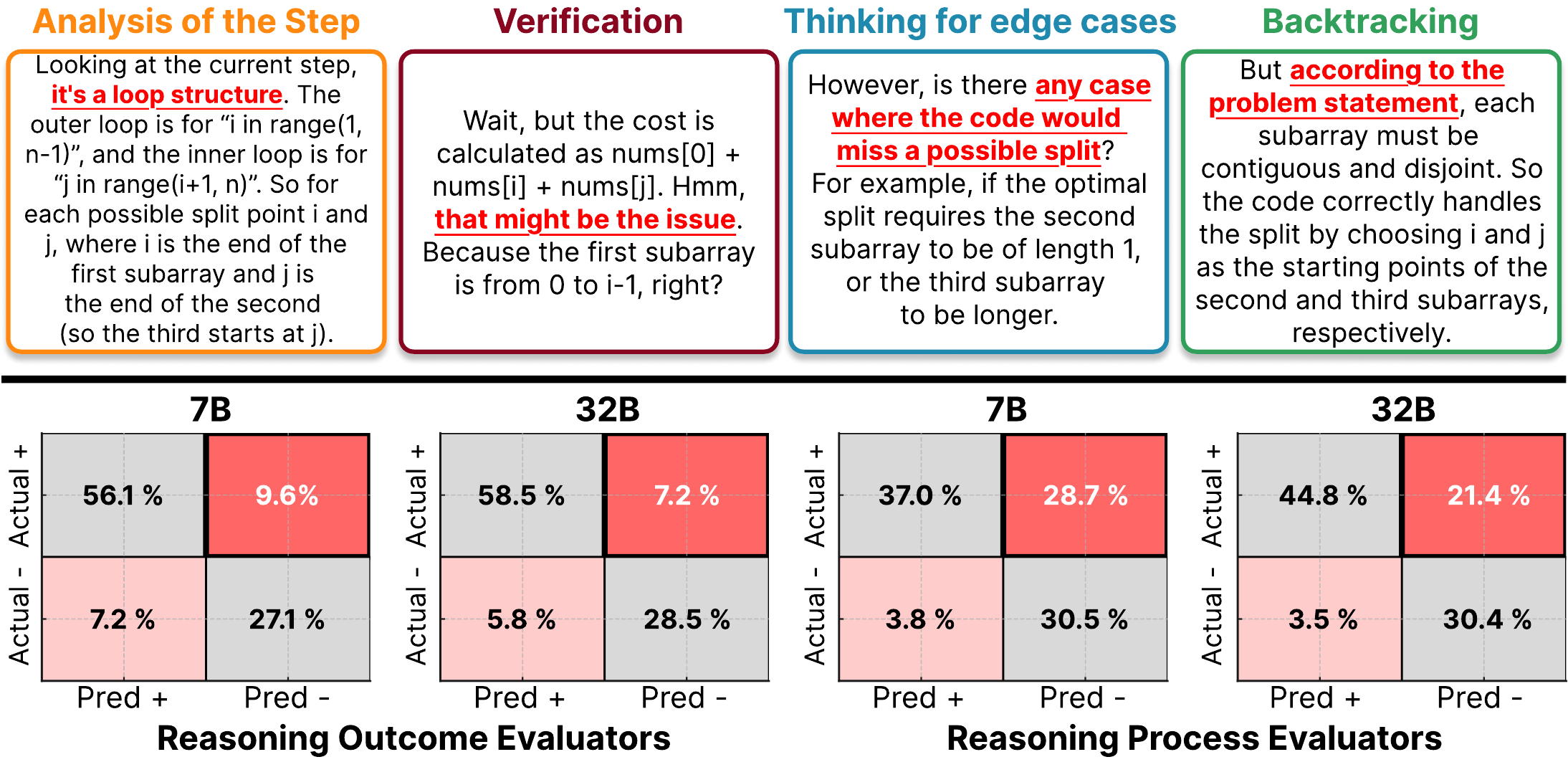}
    \caption{\textbf{(Top)} Example of reasoning process evaluator-generated CoT, which includes diverse reasoning patterns that contribute to improved evaluation. \textbf{(Bottom)} Trade-off between precision and recall in reasoning process and outcome evaluators.}
    
    \label{fig:analyses_figure}
    \vspace{-3mm}
\end{figure*}

\subsection{Experimental results}


\textbf{Finding 1: Scaling evaluation-time compute is more effective than generating additional candidate responses given a fixed compute budget}. From the top left and top middle plots in \autoref{fig:bon1}, we observe that reasoning evaluators using Best-of-8 -- including both process and outcome evaluators -- achieve performance equal to or better than direct evaluators using Best-of-64, while requiring similar or less test-time compute. Looking at the overall scaling trends, we find that reasoning evaluators suffer less from \textit{reward model over-optimization}, a phenomenon where Best-of-$N$ gains plateau or diminish with increased $N$~\citep{gao2023scaling,rafailov2024scaling}. Reward model over-optimization arises from imperfections in the reward function’s approximation of the ground truth~\citep{lambert2023alignment}; our results suggest that scaling evaluation-time compute can partially mitigate this issue by providing more robust evaluation. Similar trends are also observed in smaller evaluators (7B): see \autoref{fig:bon2} in \autoref{appendix:bon_extended}.



\textbf{Finding 2: Combining scores from reasoning outcome and process evaluation can boost performance}. From the top right plot of \autoref{fig:bon1}, we observe that combining outcome and process scores can yield improved results. We hypothesize that the two evaluation approaches provide \textit{complementary} signals that enable more accurate assessment. We investigate this hypothesis in \autoref{appendix:prcurve} and also further study the effects of altering $\alpha$ in \autoref{appendix:why}.


\textbf{Finding 3: Reasoning evaluators are especially effective for assessing code outputs}. From the bottom left and bottom middle plot in \autoref{fig:bon1}, we observe that reasoning evaluators not only significantly outperform direct evaluators while using less test-time compute, they also exhibit promising scaling trends when assessing code outputs. On the other hand, both the direct outcome evaluator and direct process evaluator heavily suffer from reward model over-optimization. We attribute these findings to two main causes. Firstly, direct process evaluators are often only trained on math data (\textit{e.g.}, PRM800K~\citep{lightman2024let}), making them less effective on out-of-domain tasks such as coding. Secondly, the heuristic-based splitting methods (\textit{e.g.}, splitting based on newline characters) typically adopted for direct process evaluation may be suboptimal for code outputs.

\textbf{Additional experiments in Appendix}. Due to space constraints, in \autoref{appendix:bon_extended}, we discuss (1) the scaling trends of 7B-sized evaluators, (2) how single-step process evaluators and fine-tuned reasoning evaluators~\citep{ankner2024critique,kim2024prometheus} compare to our baselines, and (3) the effects of applying model-based splitting and different aggregation functions. Furthermore, we explore evaluating reasoning model-generated traces using reasoning models (self-evaluation) in \autoref{appendix:self_eval}.

\section{Analyses on why reasoning process evaluators are effective}\label{sec:discussion}

\textbf{Why are reasoning evaluators effective?} The top row of \autoref{fig:analyses_figure} presents descriptions of common patterns observed in the multi-step process evaluation outputs of reasoning evaluators, accompanied by representative examples from evaluations on a LeetCode problem. 
It is known that reasoning models use versatile reasoning patterns when solving problems~\citep{gandhi2025cognitive}, such as generating ``Wait'' tokens that encourage verification between solution steps; similar patterns are also observed when reasoning models are prompted to function as evaluators. Some patterns we observe in our example include (1) examining what implementation the given step contains, (2) conducting a form of meta-verification by re-verifying their evaluation content, (3) exploring whether there are edge cases that have yet to be  identified, and (4) performing backtracking by reviewing the initial problem conditions, the content of the response, and its own evaluation process. We speculate that the ability to apply strategies used for problem-solving to the evaluation process is the key reason why reasoning models show better evaluation performance.

\textbf{When is process evaluation effective?} The confusion matrix in the bottom row of \autoref{fig:analyses_figure} illustrates the strengths and weaknesses of outcome evaluation and process evaluation. Across different model sizes, reasoning process evaluators achieve higher precision but lower recall compared to reasoning outcome evaluators. Consequently, when a reasoning process evaluator predicts that all steps are correct, the final answer is likely to be correct (false positive rates are 3.8\% and 3.5\%).

\section{Conclusion}
In this paper, we study test-time scaling applied to evaluators (evaluation-time scaling). Analogous to how test-time scaling improves generator performance, we find that evaluation-time scaling effectively enhances evaluation performance. Moreover, evaluation-time scaling can be combined with generator test-time scaling to further boost generation quality. Lastly, we show that reasoning process evaluation tends to make more conservative predictions than reasoning outcome evaluation, and that combining the two yields performance gains.

\section*{Limitations}
\label{appendix:limitations}

In this section, we provide rationales behind the datasets and models used in this study.

\textbf{Dataset selection} In \autoref{section:scaling}, we only use ProcessBench \citep{zheng2024processbench} for evaluating different evaluator implementations (\autoref{tab:processbench}). ProcessBench provides high-quality human-annotated labels obtained from diverse reasoning traces across 4 benchmarks (GSM8k, MATH, OlympiadBench, and OmniMath) and 12 generators (Llama and Qwen family with varying sizes), suitable for testing different evaluator implementations. While there are other datasets with similar purposes, we resort to ProcessBench for the following reasons:

\begin{itemize}[]
    \item Datasets like PRM800k \citep{lightman2024let}, MR-GSM8k \citep{zeng2023mr}, and MR-MATH \citep{xia2025evaluating} annotate errors from GSM8k and MATH, which is already covered by ProcessBench.
    \item While PRMBench \citep{song2025prmbench} introduces diverse error types in the MATH dataset, the dataset is generated by synthetically perturbing responses based on a manually developed taxonomy of reasoning errors. Therefore, we make use of ProcessBench that annotates naturally occurring errors in LLM-generated responses without any modification, which better suits the research question \textit{"Can we leverage evaluation-time scaling to improve the generator's performance?"}.
    \item REVEAL \citep{jacovi2024chain} annotates errors in LLM responses in commonsense reasoning benchmarks, which are not covered by ProcessBench. However, the inter-annotator agreement is significantly lower than ProcessBench ($\kappa \sim 0.5$), indicating high variance in the data label.
\end{itemize}

\textbf{Model selection} We provide the full list of models in \autoref{appendix:model_list_scaling} and \autoref{appendix:model_list_bon}. While this paper includes results from a large variety of critic models and direct evaluators, we could not test on other highly capable models due to their recency \citep{gemmateam2025gemma3technicalreport, abdin2025phi4reasoningtechnicalreport}, excessive hardware requirements (using $\sim$70B-sized models as reasoning evaluators), and limited budget for closed-source APIs (\textit{e.g.}, OpenAI o1, Gemini 2.5, Claude 3).

\textbf{Task Coverage} Another limitation of our work is that we do not assess reasoning evaluators for tasks outside math and code. We focus on math and code because (1) they are tasks that admit easy verification and (2) they align with the main strengths of reasoning models. There nonetheless exist other interesting and relevant tasks, including tasks with non-verifiable outputs such as creative and scientific writing. We encourage the research community to extend our work to include such tasks.

\section*{Acknowledgements}
\label{appendix:ack}

This work was supported in part by the National Science Foundation under Grant Nos. DMS-2502281.

\bibliography{custom}

@inproceedings{mondorf2024beyond,
title={Beyond Accuracy: Evaluating the Reasoning Behavior of Large Language Models - A Survey},
author={Philipp Mondorf and Barbara Plank},
booktitle={First Conference on Language Modeling},
year={2024},
url={https://openreview.net/forum?id=Lmjgl2n11u}
}

@article{uesato2022solving,
  title={Solving math word problems with process-and outcome-based feedback},
  author={Uesato, Jonathan and Kushman, Nate and Kumar, Ramana and Song, Francis and Siegel, Noah and Wang, Lisa and Creswell, Antonia and Irving, Geoffrey and Higgins, Irina},
  journal={arXiv preprint arXiv:2211.14275},
  year={2022}
}

@article{
liang2023holistic,
title={Holistic Evaluation of Language Models},
author={Percy Liang and Rishi Bommasani and Tony Lee and Dimitris Tsipras and Dilara Soylu and Michihiro Yasunaga and Yian Zhang and Deepak Narayanan and Yuhuai Wu and Ananya Kumar and Benjamin Newman and Binhang Yuan and Bobby Yan and Ce Zhang and Christian Alexander Cosgrove and Christopher D Manning and Christopher Re and Diana Acosta-Navas and Drew Arad Hudson and Eric Zelikman and Esin Durmus and Faisal Ladhak and Frieda Rong and Hongyu Ren and Huaxiu Yao and Jue WANG and Keshav Santhanam and Laurel Orr and Lucia Zheng and Mert Yuksekgonul and Mirac Suzgun and Nathan Kim and Neel Guha and Niladri S. Chatterji and Omar Khattab and Peter Henderson and Qian Huang and Ryan Andrew Chi and Sang Michael Xie and Shibani Santurkar and Surya Ganguli and Tatsunori Hashimoto and Thomas Icard and Tianyi Zhang and Vishrav Chaudhary and William Wang and Xuechen Li and Yifan Mai and Yuhui Zhang and Yuta Koreeda},
journal={Transactions on Machine Learning Research},
issn={2835-8856},
year={2023},
url={https://openreview.net/forum?id=iO4LZibEqW},
note={Featured Certification, Expert Certification}
}

@inproceedings{
moskovitz2024confronting,
title={Confronting Reward Model Overoptimization with Constrained {RLHF}},
author={Ted Moskovitz and Aaditya K Singh and DJ Strouse and Tuomas Sandholm and Ruslan Salakhutdinov and Anca Dragan and Stephen Marcus McAleer},
booktitle={The Twelfth International Conference on Learning Representations},
year={2024},
url={https://openreview.net/forum?id=gkfUvn0fLU}
}

@article{huang202237,
  title={The 37 implementation details of proximal policy optimization},
  author={Huang, Shengyi and Dossa, Rousslan Fernand Julien and Raffin, Antonin and Kanervisto, Anssi and Wang, Weixun},
  journal={The ICLR Blog Track 2023},
  year={2022}
}

@inproceedings{
rafailov2024scaling,
title={Scaling Laws for Reward Model Overoptimization in Direct Alignment Algorithms},
author={Rafael Rafailov and Yaswanth Chittepu and Ryan Park and Harshit Sikchi and Joey Hejna and W. Bradley Knox and Chelsea Finn and Scott Niekum},
booktitle={The Thirty-eighth Annual Conference on Neural Information Processing Systems},
year={2024},
url={https://openreview.net/forum?id=pf4OuJyn4Q}
}

@article{stiennon2020learning,
  title={Learning to summarize with human feedback},
  author={Stiennon, Nisan and Ouyang, Long and Wu, Jeffrey and Ziegler, Daniel and Lowe, Ryan and Voss, Chelsea and Radford, Alec and Amodei, Dario and Christiano, Paul F},
  journal={Advances in neural information processing systems},
  volume={33},
  pages={3008--3021},
  year={2020}
}

@inproceedings{coste2024reward,
title={Reward Model Ensembles Help Mitigate Overoptimization},
author={Thomas Coste and Usman Anwar and Robert Kirk and David Krueger},
booktitle={The Twelfth International Conference on Learning Representations},
year={2024},
url={https://openreview.net/forum?id=dcjtMYkpXx}
}

@inproceedings{gao2023scaling,
  title={Scaling laws for reward model overoptimization},
  author={Gao, Leo and Schulman, John and Hilton, Jacob},
  booktitle={International Conference on Machine Learning},
  pages={10835--10866},
  year={2023},
  organization={PMLR}
}

@article{lambert2024rewardbench,
  title={Rewardbench: Evaluating reward models for language modeling},
  author={Lambert, Nathan and Pyatkin, Valentina and Morrison, Jacob and Miranda, LJ and Lin, Bill Yuchen and Chandu, Khyathi and Dziri, Nouha and Kumar, Sachin and Zick, Tom and Choi, Yejin and others},
  journal={arXiv preprint arXiv:2403.13787},
  year={2024}
}

@article{bai2022training,
  title={Training a helpful and harmless assistant with reinforcement learning from human feedback},
  author={Bai, Yuntao and Jones, Andy and Ndousse, Kamal and Askell, Amanda and Chen, Anna and DasSarma, Nova and Drain, Dawn and Fort, Stanislav and Ganguli, Deep and Henighan, Tom and others},
  journal={arXiv preprint arXiv:2204.05862},
  year={2022}
}

@article{ouyang2022training,
  title={Training language models to follow instructions with human feedback},
  author={Ouyang, Long and Wu, Jeffrey and Jiang, Xu and Almeida, Diogo and Wainwright, Carroll and Mishkin, Pamela and Zhang, Chong and Agarwal, Sandhini and Slama, Katarina and Ray, Alex and others},
  journal={Advances in neural information processing systems},
  volume={35},
  pages={27730--27744},
  year={2022}
}

@inproceedings{
    kim2024prometheus,
    title={Prometheus: Inducing Evaluation Capability in Language Models},
    author={Seungone Kim and Jamin Shin and Yejin Cho and Joel Jang and Shayne Longpre and Hwaran Lee and Sangdoo Yun and Seongjin Shin and Sungdong Kim and James Thorne and Minjoon Seo},
    booktitle={The Twelfth International Conference on Learning Representations},
    year={2024},
    url={https://openreview.net/forum?id=8euJaTveKw}
}

@article{kim2024prometheus2,
  title={Prometheus 2: An Open Source Language Model Specialized in Evaluating Other Language Models},
  author={Kim, Seungone and Suk, Juyoung and Longpre, Shayne and Lin, Bill Yuchen and Shin, Jamin and Welleck, Sean and Neubig, Graham and Lee, Moontae and Lee, Kyungjae and Seo, Minjoon},
  journal={arXiv preprint arXiv:2405.01535},
  year={2024}
}

@article{zheng2024judging,
  title={Judging llm-as-a-judge with mt-bench and chatbot arena},
  author={Zheng, Lianmin and Chiang, Wei-Lin and Sheng, Ying and Zhuang, Siyuan and Wu, Zhanghao and Zhuang, Yonghao and Lin, Zi and Li, Zhuohan and Li, Dacheng and Xing, Eric and others},
  journal={Advances in Neural Information Processing Systems},
  volume={36},
  year={2024}
}

@inproceedings{
    ye2024flask,
    title={{FLASK}: Fine-grained Language Model Evaluation based on Alignment Skill Sets},
    author={Seonghyeon Ye and Doyoung Kim and Sungdong Kim and Hyeonbin Hwang and Seungone Kim and Yongrae Jo and James Thorne and Juho Kim and Minjoon Seo},
    booktitle={The Twelfth International Conference on Learning Representations},
    year={2024},
    url={https://openreview.net/forum?id=CYmF38ysDa}
}

@article{dubois2024length,
  title={Length-controlled alpacaeval: A simple way to debias automatic evaluators},
  author={Dubois, Yann and Galambosi, Bal{\'a}zs and Liang, Percy and Hashimoto, Tatsunori B},
  journal={arXiv preprint arXiv:2404.04475},
  year={2024}
}

@misc{arenahard2024,
    title = {From Live Data to High-Quality Benchmarks: The Arena-Hard Pipeline},
    url = {https://lmsys.org/blog/2024-04-19-arena-hard/},
    author = {Tianle Li and Wei-Lin Chiang and Evan Frick and Lisa Dunlap and Banghua Zhu and Joseph E. Gonzalez and Ion Stoica},
    month = {April},
    year = {2024}
}

@article{dong2024rlhf,
  title={RLHF Workflow: From Reward Modeling to Online RLHF},
  author={Dong, Hanze and Xiong, Wei and Pang, Bo and Wang, Haoxiang and Zhao, Han and Zhou, Yingbo and Jiang, Nan and Sahoo, Doyen and Xiong, Caiming and Zhang, Tong},
  journal={arXiv preprint arXiv:2405.07863},
  year={2024}
}

@misc{dubey2024llama3herdmodels,
  title =         {The Llama 3 Herd of Models},
  author =        {Llama Team, AI @ Meta},
  year =          {2024},
  eprint =        {2407.21783},
  archivePrefix = {arXiv},
  primaryClass =  {cs.AI},
  url =           {https://arxiv.org/abs/2407.21783}
}

@inproceedings{longpre2023flan,
  title={The flan collection: Designing data and methods for effective instruction tuning},
  author={Longpre, Shayne and Hou, Le and Vu, Tu and Webson, Albert and Chung, Hyung Won and Tay, Yi and Zhou, Denny and Le, Quoc V and Zoph, Barret and Wei, Jason and others},
  booktitle={International Conference on Machine Learning},
  pages={22631--22648},
  year={2023},
  organization={PMLR}
}

@article{zheng2024processbench,
  title={Processbench: Identifying process errors in mathematical reasoning},
  author={Zheng, Chujie and Zhang, Zhenru and Zhang, Beichen and Lin, Runji and Lu, Keming and Yu, Bowen and Liu, Dayiheng and Zhou, Jingren and Lin, Junyang},
  journal={arXiv preprint arXiv:2412.06559},
  year={2024}
}

@article{snell2024scaling,
  title={Scaling llm test-time compute optimally can be more effective than scaling model parameters},
  author={Snell, Charlie and Lee, Jaehoon and Xu, Kelvin and Kumar, Aviral},
  journal={arXiv preprint arXiv:2408.03314},
  year={2024}
}

@article{zhang2025lessons,
  title={The Lessons of Developing Process Reward Models in Mathematical Reasoning},
  author={Zhang, Zhenru and Zheng, Chujie and Wu, Yangzhen and Zhang, Beichen and Lin, Runji and Yu, Bowen and Liu, Dayiheng and Zhou, Jingren and Lin, Junyang},
  journal={arXiv preprint arXiv:2501.07301},
  year={2025}
}

@inproceedings{wang2024math,
  title={Math-shepherd: Verify and reinforce llms step-by-step without human annotations},
  author={Wang, Peiyi and Li, Lei and Shao, Zhihong and Xu, Runxin and Dai, Damai and Li, Yifei and Chen, Deli and Wu, Yu and Sui, Zhifang},
  booktitle={Proceedings of the 62nd Annual Meeting of the Association for Computational Linguistics (Volume 1: Long Papers)},
  pages={9426--9439},
  year={2024}
}

@article{liu2024skywork,
  title={Skywork-reward: Bag of tricks for reward modeling in llms},
  author={Liu, Chris Yuhao and Zeng, Liang and Liu, Jiacai and Yan, Rui and He, Jujie and Wang, Chaojie and Yan, Shuicheng and Liu, Yang and Zhou, Yahui},
  journal={arXiv preprint arXiv:2410.18451},
  year={2024}
}

@article{cobbe2021training,
  title={Training verifiers to solve math word problems},
  author={Cobbe, Karl and Kosaraju, Vineet and Bavarian, Mohammad and Chen, Mark and Jun, Heewoo and Kaiser, Lukasz and Plappert, Matthias and Tworek, Jerry and Hilton, Jacob and Nakano, Reiichiro and others},
  journal={arXiv preprint arXiv:2110.14168},
  year={2021}
}

@inproceedings{lightman2024let,
  title={Let's Verify Step by Step},
  author={Lightman, Hunter and Kosaraju, Vineet and Burda, Yuri and Edwards, Harrison and Baker, Bowen and Lee, Teddy and Leike, Jan and Schulman, John and Sutskever, Ilya and Cobbe, Karl},
  booktitle={The Twelfth International Conference on Learning Representations},
  year={2024}
}

@inproceedings{kim2023prometheus,
  title={Prometheus: Inducing fine-grained evaluation capability in language models},
  author={Kim, Seungone and Shin, Jamin and Cho, Yejin and Jang, Joel and Longpre, Shayne and Lee, Hwaran and Yun, Sangdoo and Shin, Seongjin and Kim, Sungdong and Thorne, James and others},
  booktitle={The Twelfth International Conference on Learning Representations},
  year={2023}
}

@article{xu2025towards,
  title={Towards Large Reasoning Models: A Survey of Reinforced Reasoning with Large Language Models},
  author={Xu, Fengli and Hao, Qianyue and Zong, Zefang and Wang, Jingwei and Zhang, Yunke and Wang, Jingyi and Lan, Xiaochong and Gong, Jiahui and Ouyang, Tianjian and Meng, Fanjin and others},
  journal={arXiv preprint arXiv:2501.09686},
  year={2025}
}

@article{jaech2024openai,
  title={Openai o1 system card},
  author={Jaech, Aaron and Kalai, Adam and Lerer, Adam and Richardson, Adam and El-Kishky, Ahmed and Low, Aiden and Helyar, Alec and Madry, Aleksander and Beutel, Alex and Carney, Alex and others},
  journal={arXiv preprint arXiv:2412.16720},
  year={2024}
}

@inproceedings{kim2023cot,
  title={The CoT Collection: Improving Zero-shot and Few-shot Learning of Language Models via Chain-of-Thought Fine-Tuning},
  author={Kim, Seungone and Joo, Se and Kim, Doyoung and Jang, Joel and Ye, Seonghyeon and Shin, Jamin and Seo, Minjoon},
  booktitle={Proceedings of the 2023 Conference on Empirical Methods in Natural Language Processing},
  pages={12685--12708},
  year={2023}
}

@article{aggarwal2025l1,
  title={L1: Controlling How Long A Reasoning Model Thinks With Reinforcement Learning},
  author={Aggarwal, Pranjal and Welleck, Sean},
  journal={arXiv preprint arXiv:2503.04697},
  year={2025}
}

@inproceedings{ankner2024critique,
  title={Critique-out-Loud Reward Models},
  author={Ankner, Zachary and Paul, Mansheej and Cui, Brandon and Chang, Jonathan Daniel and Ammanabrolu, Prithviraj},
  booktitle={Pluralistic Alignment Workshop at NeurIPS 2024},
  year={2024}
}

@inproceedings{
wang2023selfconsistency,
title={Self-Consistency Improves Chain of Thought Reasoning in Language Models},
author={Xuezhi Wang and Jason Wei and Dale Schuurmans and Quoc V Le and Ed H. Chi and Sharan Narang and Aakanksha Chowdhery and Denny Zhou},
booktitle={The Eleventh International Conference on Learning Representations },
year={2023},
url={https://openreview.net/forum?id=1PL1NIMMrw}
}

@misc{qwen2.5,
    title = {Qwen2.5: A Party of Foundation Models},
    url = {https://qwenlm.github.io/blog/qwen2.5/},
    author = {Qwen Team},
    month = {September},
    year = {2024}
}

@misc{qwq-32b-preview,
    title = {QwQ: Reflect Deeply on the Boundaries of the Unknown},
    url = {https://qwenlm.github.io/blog/qwq-32b-preview/},
    author = {Qwen Team},
    month = {November},
    year = {2024}
}

@misc{xiong2024rlhflowmath,
      author={Wei Xiong and Hanning Zhang and Nan Jiang and Tong Zhang},
  title = {An Implementation of Generative PRM},
  year = {2024},
  publisher = {GitHub},
  journal = {GitHub repository},
  howpublished = {\url{https://github.com/RLHFlow/RLHF-Reward-Modeling}}
}

@misc{skyworkopeno12024,
  title={Skywork-o1 Open Series},
  author={Skywork-o1 Team},
  year={2024},
  month={November},
  howpublished={\url{https://huggingface.co/Skywork}},
  url={https://huggingface.co/Skywork},
}

@inproceedings{
villalobos2024position,
title={Position: Will we run out of data? Limits of {LLM} scaling based on human-generated data},
author={Pablo Villalobos and Anson Ho and Jaime Sevilla and Tamay Besiroglu and Lennart Heim and Marius Hobbhahn},
booktitle={Forty-first International Conference on Machine Learning},
year={2024},
url={https://openreview.net/forum?id=ViZcgDQjyG}
}

@article{hoffmann2022empirical,
  title={An empirical analysis of compute-optimal large language model training},
  author={Hoffmann, Jordan and Borgeaud, Sebastian and Mensch, Arthur and Buchatskaya, Elena and Cai, Trevor and Rutherford, Eliza and de Las Casas, Diego and Hendricks, Lisa Anne and Welbl, Johannes and Clark, Aidan and others},
  journal={Advances in neural information processing systems},
  volume={35},
  pages={30016--30030},
  year={2022}
}

@article{kaplan2020scaling,
  title={Scaling laws for neural language models},
  author={Kaplan, Jared and McCandlish, Sam and Henighan, Tom and Brown, Tom B and Chess, Benjamin and Child, Rewon and Gray, Scott and Radford, Alec and Wu, Jeffrey and Amodei, Dario},
  journal={arXiv preprint arXiv:2001.08361},
  year={2020}
}

@inproceedings{
yeo2025demystifying,
title={Demystifying Long Chain-of-Thought Reasoning in {LLM}s},
author={Edward Yeo and Yuxuan Tong and Xinyao Niu and Graham Neubig and Xiang Yue},
booktitle={Scaling Self-Improving Foundation Models without Human Supervision},
year={2025},
url={https://openreview.net/forum?id=6A861u4Crm}
}

@article{muennighoff2025s1,
  title={s1: Simple test-time scaling},
  author={Muennighoff, Niklas and Yang, Zitong and Shi, Weijia and Li, Xiang Lisa and Fei-Fei, Li and Hajishirzi, Hannaneh and Zettlemoyer, Luke and Liang, Percy and Cand{\`e}s, Emmanuel and Hashimoto, Tatsunori},
  journal={arXiv preprint arXiv:2501.19393},
  year={2025}
}

@article{welleck2024from,
title={From Decoding to Meta-Generation: Inference-time Algorithms for Large Language Models},
author={Sean Welleck and Amanda Bertsch and Matthew Finlayson and Hailey Schoelkopf and Alex Xie and Graham Neubig and Ilia Kulikov and Zaid Harchaoui},
journal={Transactions on Machine Learning Research},
issn={2835-8856},
year={2024},
url={https://openreview.net/forum?id=eskQMcIbMS},
note={Survey Certification}
}

@article{guo2025deepseek,
  title={Deepseek-r1: Incentivizing reasoning capability in llms via reinforcement learning},
  author={Guo, Daya and Yang, Dejian and Zhang, Haowei and Song, Junxiao and Zhang, Ruoyu and Xu, Runxin and Zhu, Qihao and Ma, Shirong and Wang, Peiyi and Bi, Xiao and others},
  journal={arXiv preprint arXiv:2501.12948},
  year={2025}
}

@inproceedings{sun2024easytohard,
title={Easy-to-Hard Generalization: Scalable Alignment Beyond Human Supervision},
author={Zhiqing Sun and Longhui Yu and Yikang Shen and Weiyang Liu and Yiming Yang and Sean Welleck and Chuang Gan},
booktitle={The Thirty-eighth Annual Conference on Neural Information Processing Systems},
year={2024},
url={https://openreview.net/forum?id=qwgfh2fTtN}
}

@article{cui2025process,
  title={Process reinforcement through implicit rewards},
  author={Cui, Ganqu and Yuan, Lifan and Wang, Zefan and Wang, Hanbin and Li, Wendi and He, Bingxiang and Fan, Yuchen and Yu, Tianyu and Xu, Qixin and Chen, Weize and others},
  journal={arXiv preprint arXiv:2502.01456},
  year={2025}
}

@inproceedings{rein2024gpqa,
title={{GPQA}: A Graduate-Level Google-Proof Q\&A Benchmark},
author={David Rein and Betty Li Hou and Asa Cooper Stickland and Jackson Petty and Richard Yuanzhe Pang and Julien Dirani and Julian Michael and Samuel R. Bowman},
booktitle={First Conference on Language Modeling},
year={2024},
url={https://openreview.net/forum?id=Ti67584b98}
}

@article{guo2024deepseek,
  title={DeepSeek-Coder: When the Large Language Model Meets Programming--The Rise of Code Intelligence},
  author={Guo, Daya and Zhu, Qihao and Yang, Dejian and Xie, Zhenda and Dong, Kai and Zhang, Wentao and Chen, Guanting and Bi, Xiao and Li, YK and others},
  journal={arXiv preprint arXiv:2401.14196},
  year={2024}
}

@inproceedings{hendrycks2021measuring,
title={Measuring Mathematical Problem Solving With the {MATH} Dataset},
author={Dan Hendrycks and Collin Burns and Saurav Kadavath and Akul Arora and Steven Basart and Eric Tang and Dawn Song and Jacob Steinhardt},
booktitle={Thirty-fifth Conference on Neural Information Processing Systems Datasets and Benchmarks Track (Round 2)},
year={2021},
url={https://openreview.net/forum?id=7Bywt2mQsCe}
}

@inproceedings{he2024olympiadbench,
  title={OlympiadBench: A Challenging Benchmark for Promoting AGI with Olympiad-Level Bilingual Multimodal Scientific Problems},
  author={He, Chaoqun and Luo, Renjie and Bai, Yuzhuo and Hu, Shengding and Thai, Zhen and Shen, Junhao and Hu, Jinyi and Han, Xu and Huang, Yujie and Zhang, Yuxiang and others},
  booktitle={Proceedings of the 62nd Annual Meeting of the Association for Computational Linguistics (Volume 1: Long Papers)},
  pages={3828--3850},
  year={2024}
}

@article{lewkowycz2022solving,
  title={Solving quantitative reasoning problems with language models},
  author={Lewkowycz, Aitor and Andreassen, Anders and Dohan, David and Dyer, Ethan and Michalewski, Henryk and Ramasesh, Vinay and Slone, Ambrose and Anil, Cem and Schlag, Imanol and Gutman-Solo, Theo and others},
  journal={Advances in Neural Information Processing Systems},
  volume={35},
  pages={3843--3857},
  year={2022}
}

@article{yang2024qwen2,
  title={Qwen2. 5-math technical report: Toward mathematical expert model via self-improvement},
  author={Yang, An and Zhang, Beichen and Hui, Binyuan and Gao, Bofei and Yu, Bowen and Li, Chengpeng and Liu, Dayiheng and Tu, Jianhong and Zhou, Jingren and Lin, Junyang and others},
  journal={arXiv preprint arXiv:2409.12122},
  year={2024}
}

@article{kim2024biggen,
  title={The biggen bench: A principled benchmark for fine-grained evaluation of language models with language models},
  author={Kim, Seungone and Suk, Juyoung and Cho, Ji Yong and Longpre, Shayne and Kim, Chaeeun and Yoon, Dongkeun and Son, Guijin and Cho, Yejin and Shafayat, Sheikh and Baek, Jinheon and others},
  journal={arXiv preprint arXiv:2406.05761},
  year={2024}
}

@inproceedings{zhang2024generative,
title={Generative Verifiers: Reward Modeling as Next-Token Prediction},
author={Lunjun Zhang and Arian Hosseini and Hritik Bansal and Mehran Kazemi and Aviral Kumar and Rishabh Agarwal},
booktitle={The 4th Workshop on Mathematical Reasoning and AI at NeurIPS'24},
year={2024},
url={https://openreview.net/forum?id=CxHRoTLmPX}
}

@inproceedings{wu2024scaling,
title={Scaling Inference Computation: Compute-Optimal Inference for Problem-Solving with Language Models},
author={Yangzhen Wu and Zhiqing Sun and Shanda Li and Sean Welleck and Yiming Yang},
booktitle={The 4th Workshop on Mathematical Reasoning and AI at NeurIPS'24},
year={2024},
url={https://openreview.net/forum?id=j7DZWSc8qu}
}

@article{gu2024survey,
  title={A Survey on LLM-as-a-Judge},
  author={Gu, Jiawei and Jiang, Xuhui and Shi, Zhichao and Tan, Hexiang and Zhai, Xuehao and Xu, Chengjin and Li, Wei and Shen, Yinghan and Ma, Shengjie and Liu, Honghao and others},
  journal={arXiv preprint arXiv:2411.15594},
  year={2024}
}

@article{wu2024better,
  title={Better instruction-following through minimum bayes risk},
  author={Wu, Ian and Fernandes, Patrick and Bertsch, Amanda and Kim, Seungone and Pakazad, Sina and Neubig, Graham},
  journal={arXiv preprint arXiv:2410.02902},
  year={2024}
}

@article{lee2025evaluating,
  title={Evaluating Step-by-step Reasoning Traces: A Survey},
  author={Lee, Jinu and Hockenmaier, Julia},
  journal={arXiv preprint arXiv:2502.12289},
  year={2025}
}

@article{lambert2023alignment,
  title={The alignment ceiling: Objective mismatch in reinforcement learning from human feedback},
  author={Lambert, Nathan and Calandra, Roberto},
  journal={arXiv preprint arXiv:2311.00168},
  year={2023}
}

@article{wang2025examining,
  title={Examining False Positives under Inference Scaling for Mathematical Reasoning},
  author={Wang, Yu and Yang, Nan and Wang, Liang and Wei, Furu},
  journal={arXiv preprint arXiv:2502.06217},
  year={2025}
}

@article{kalra2025verdict,
  title={Verdict: A Library for Scaling Judge-Time Compute},
  author={Kalra, Nimit and Tang, Leonard},
  journal={arXiv preprint arXiv:2502.18018},
  year={2025}
}

@article{gandhi2025cognitive,
  title={Cognitive behaviors that enable self-improving reasoners, or, four habits of highly effective stars},
  author={Gandhi, Kanishk and Chakravarthy, Ayush and Singh, Anikait and Lile, Nathan and Goodman, Noah D},
  journal={arXiv preprint arXiv:2503.01307},
  year={2025}
}

@article{son2025linguistic,
  title={Linguistic Generalizability of Test-Time Scaling in Mathematical Reasoning},
  author={Son, Guijin and Hong, Jiwoo and Ko, Hyunwoo and Thorne, James},
  journal={arXiv preprint arXiv:2502.17407},
  year={2025}
}

@misc{gandhi2025cognitivebehaviorsenableselfimproving,
      title={Cognitive Behaviors that Enable Self-Improving Reasoners, or, Four Habits of Highly Effective STaRs}, 
      author={Kanishk Gandhi and Ayush Chakravarthy and Anikait Singh and Nathan Lile and Noah D. Goodman},
      year={2025},
      eprint={2503.01307},
      archivePrefix={arXiv},
      primaryClass={cs.CL},
      url={https://arxiv.org/abs/2503.01307}, 
}

@misc{lu2025retrosearchexploringuntakenpaths,
      title={Retro-Search: Exploring Untaken Paths for Deeper and Efficient Reasoning}, 
      author={Ximing Lu and Seungju Han and David Acuna and Hyunwoo Kim and Jaehun Jung and Shrimai Prabhumoye and Niklas Muennighoff and Mostofa Patwary and Mohammad Shoeybi and Bryan Catanzaro and Yejin Choi},
      year={2025},
      eprint={2504.04383},
      archivePrefix={arXiv},
      primaryClass={cs.AI},
      url={https://arxiv.org/abs/2504.04383}, 
}

@inproceedings{lyu2023faithful,
  title={Faithful chain-of-thought reasoning},
  author={Lyu, Qing and Havaldar, Shreya and Stein, Adam and Zhang, Li and Rao, Delip and Wong, Eric and Apidianaki, Marianna and Callison-Burch, Chris},
  booktitle={The 13th International Joint Conference on Natural Language Processing and the 3rd Conference of the Asia-Pacific Chapter of the Association for Computational Linguistics (IJCNLP-AACL 2023)},
  year={2023}
}

@article{zeng2023mr,
  title={Mr-gsm8k: A meta-reasoning benchmark for large language model evaluation},
  author={Zeng, Zhongshen and Chen, Pengguang and Liu, Shu and Jiang, Haiyun and Jia, Jiaya},
  journal={arXiv preprint arXiv:2312.17080},
  year={2023}
}

@inproceedings{xia2025evaluating,
  title={Evaluating mathematical reasoning beyond accuracy},
  author={Xia, Shijie and Li, Xuefeng and Liu, Yixin and Wu, Tongshuang and Liu, Pengfei},
  booktitle={Proceedings of the AAAI Conference on Artificial Intelligence},
  volume={39},
  number={26},
  pages={27723--27730},
  year={2025}
}

@article{jacovi2024chain,
  title={A chain-of-thought is as strong as its weakest link: A benchmark for verifiers of reasoning chains},
  author={Jacovi, Alon and Bitton, Yonatan and Bohnet, Bernd and Herzig, Jonathan and Honovich, Or and Tseng, Michael and Collins, Michael and Aharoni, Roee and Geva, Mor},
  journal={arXiv preprint arXiv:2402.00559},
  year={2024}
}

@article{song2025prmbench,
  title={PRMBench: A Fine-grained and Challenging Benchmark for Process-Level Reward Models},
  author={Song, Mingyang and Su, Zhaochen and Qu, Xiaoye and Zhou, Jiawei and Cheng, Yu},
  journal={arXiv preprint arXiv:2501.03124},
  year={2025}
}

@misc{gemmateam2025gemma3technicalreport,
      title={Gemma 3 Technical Report}, 
      author={Gemma Team and Aishwarya Kamath and Johan Ferret and Shreya Pathak and Nino Vieillard and Ramona Merhej and Sarah Perrin and Tatiana Matejovicova and Alexandre Ramé and Morgane Rivière and Louis Rouillard and Thomas Mesnard and Geoffrey Cideron and Jean-bastien Grill and Sabela Ramos and Edouard Yvinec and Michelle Casbon and Etienne Pot and Ivo Penchev and Gaël Liu and Francesco Visin and Kathleen Kenealy and Lucas Beyer and Xiaohai Zhai and Anton Tsitsulin and Robert Busa-Fekete and Alex Feng and Noveen Sachdeva and Benjamin Coleman and Yi Gao and Basil Mustafa and Iain Barr and Emilio Parisotto and David Tian and Matan Eyal and Colin Cherry and Jan-Thorsten Peter and Danila Sinopalnikov and Surya Bhupatiraju and Rishabh Agarwal and Mehran Kazemi and Dan Malkin and Ravin Kumar and David Vilar and Idan Brusilovsky and Jiaming Luo and Andreas Steiner and Abe Friesen and Abhanshu Sharma and Abheesht Sharma and Adi Mayrav Gilady and Adrian Goedeckemeyer and Alaa Saade and Alex Feng and Alexander Kolesnikov and Alexei Bendebury and Alvin Abdagic and Amit Vadi and András György and André Susano Pinto and Anil Das and Ankur Bapna and Antoine Miech and Antoine Yang and Antonia Paterson and Ashish Shenoy and Ayan Chakrabarti and Bilal Piot and Bo Wu and Bobak Shahriari and Bryce Petrini and Charlie Chen and Charline Le Lan and Christopher A. Choquette-Choo and CJ Carey and Cormac Brick and Daniel Deutsch and Danielle Eisenbud and Dee Cattle and Derek Cheng and Dimitris Paparas and Divyashree Shivakumar Sreepathihalli and Doug Reid and Dustin Tran and Dustin Zelle and Eric Noland and Erwin Huizenga and Eugene Kharitonov and Frederick Liu and Gagik Amirkhanyan and Glenn Cameron and Hadi Hashemi and Hanna Klimczak-Plucińska and Harman Singh and Harsh Mehta and Harshal Tushar Lehri and Hussein Hazimeh and Ian Ballantyne and Idan Szpektor and Ivan Nardini and Jean Pouget-Abadie and Jetha Chan and Joe Stanton and John Wieting and Jonathan Lai and Jordi Orbay and Joseph Fernandez and Josh Newlan and Ju-yeong Ji and Jyotinder Singh and Kat Black and Kathy Yu and Kevin Hui and Kiran Vodrahalli and Klaus Greff and Linhai Qiu and Marcella Valentine and Marina Coelho and Marvin Ritter and Matt Hoffman and Matthew Watson and Mayank Chaturvedi and Michael Moynihan and Min Ma and Nabila Babar and Natasha Noy and Nathan Byrd and Nick Roy and Nikola Momchev and Nilay Chauhan and Noveen Sachdeva and Oskar Bunyan and Pankil Botarda and Paul Caron and Paul Kishan Rubenstein and Phil Culliton and Philipp Schmid and Pier Giuseppe Sessa and Pingmei Xu and Piotr Stanczyk and Pouya Tafti and Rakesh Shivanna and Renjie Wu and Renke Pan and Reza Rokni and Rob Willoughby and Rohith Vallu and Ryan Mullins and Sammy Jerome and Sara Smoot and Sertan Girgin and Shariq Iqbal and Shashir Reddy and Shruti Sheth and Siim Põder and Sijal Bhatnagar and Sindhu Raghuram Panyam and Sivan Eiger and Susan Zhang and Tianqi Liu and Trevor Yacovone and Tyler Liechty and Uday Kalra and Utku Evci and Vedant Misra and Vincent Roseberry and Vlad Feinberg and Vlad Kolesnikov and Woohyun Han and Woosuk Kwon and Xi Chen and Yinlam Chow and Yuvein Zhu and Zichuan Wei and Zoltan Egyed and Victor Cotruta and Minh Giang and Phoebe Kirk and Anand Rao and Kat Black and Nabila Babar and Jessica Lo and Erica Moreira and Luiz Gustavo Martins and Omar Sanseviero and Lucas Gonzalez and Zach Gleicher and Tris Warkentin and Vahab Mirrokni and Evan Senter and Eli Collins and Joelle Barral and Zoubin Ghahramani and Raia Hadsell and Yossi Matias and D. Sculley and Slav Petrov and Noah Fiedel and Noam Shazeer and Oriol Vinyals and Jeff Dean and Demis Hassabis and Koray Kavukcuoglu and Clement Farabet and Elena Buchatskaya and Jean-Baptiste Alayrac and Rohan Anil and Dmitry and Lepikhin and Sebastian Borgeaud and Olivier Bachem and Armand Joulin and Alek Andreev and Cassidy Hardin and Robert Dadashi and Léonard Hussenot},
      year={2025},
      eprint={2503.19786},
      archivePrefix={arXiv},
      primaryClass={cs.CL},
      url={https://arxiv.org/abs/2503.19786}, 
}

@misc{abdin2025phi4reasoningtechnicalreport,
      title={Phi-4-reasoning Technical Report}, 
      author={Marah Abdin and Sahaj Agarwal and Ahmed Awadallah and Vidhisha Balachandran and Harkirat Behl and Lingjiao Chen and Gustavo de Rosa and Suriya Gunasekar and Mojan Javaheripi and Neel Joshi and Piero Kauffmann and Yash Lara and Caio César Teodoro Mendes and Arindam Mitra and Besmira Nushi and Dimitris Papailiopoulos and Olli Saarikivi and Shital Shah and Vaishnavi Shrivastava and Vibhav Vineet and Yue Wu and Safoora Yousefi and Guoqing Zheng},
      year={2025},
      eprint={2504.21318},
      archivePrefix={arXiv},
      primaryClass={cs.AI},
      url={https://arxiv.org/abs/2504.21318}, 
}

@article{brown2024large,
  title={Large language monkeys: Scaling inference compute with repeated sampling},
  author={Brown, Bradley and Juravsky, Jordan and Ehrlich, Ryan and Clark, Ronald and Le, Quoc V and R{\'e}, Christopher and Mirhoseini, Azalia},
  journal={arXiv preprint arXiv:2407.21787},
  year={2024}
}

@article{wang2024mmlu,
  title={Mmlu-pro: A more robust and challenging multi-task language understanding benchmark},
  author={Wang, Yubo and Ma, Xueguang and Zhang, Ge and Ni, Yuansheng and Chandra, Abhranil and Guo, Shiguang and Ren, Weiming and Arulraj, Aaran and He, Xuan and Jiang, Ziyan and others},
  journal={Advances in Neural Information Processing Systems},
  volume={37},
  pages={95266--95290},
  year={2024}
}

\appendix
\newpage



\clearpage

\begin{figure*}[t]
    \centering
    \includegraphics[width=1\linewidth]{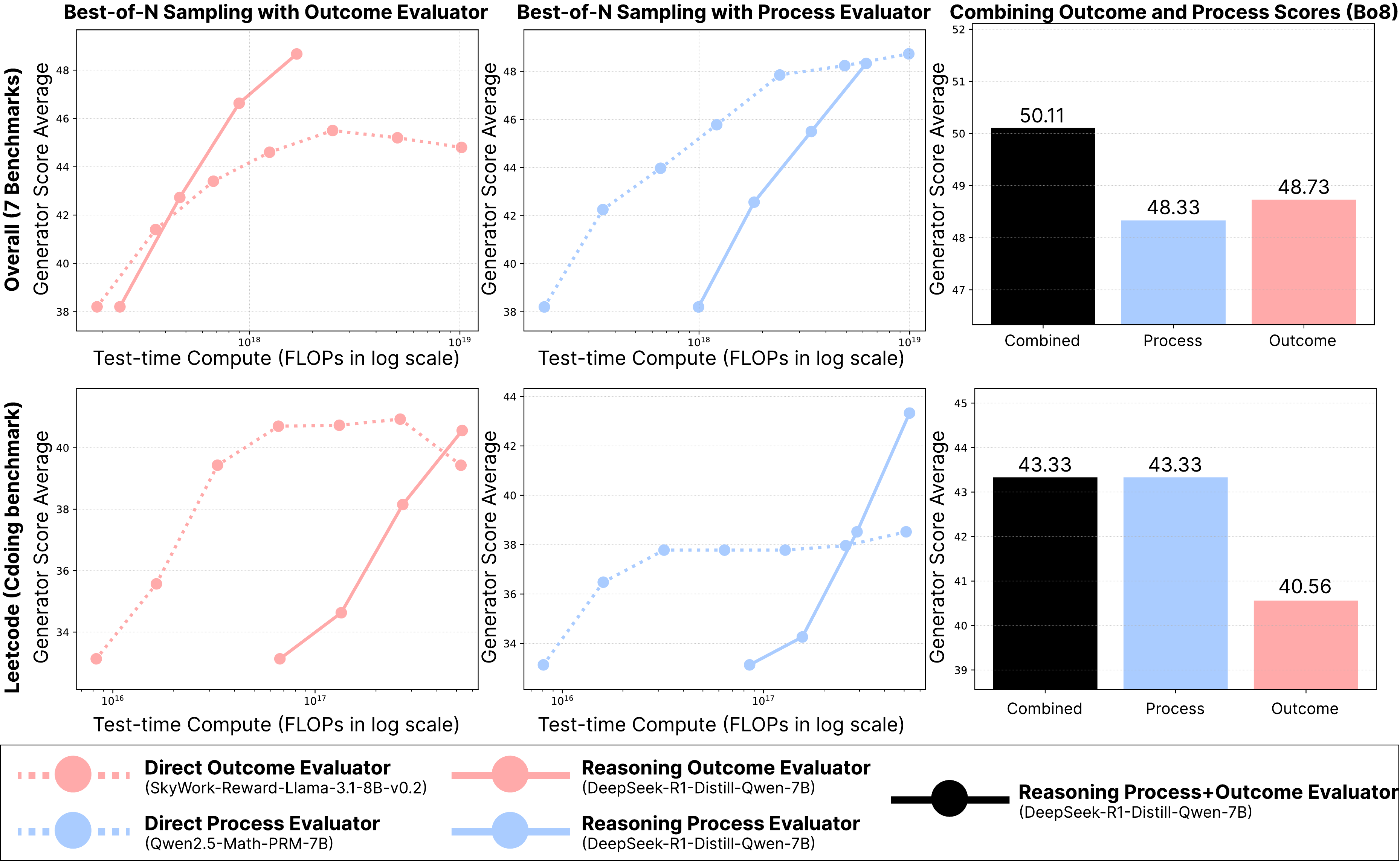}
    \caption{Best-of-$N$ experimental results for $\sim$ 7B scale evaluators. \textbf{(left, middle)} We compare direct evaluators (using Best-of-64) against reasoning evaluators (using Best-of-8), with each dot representing a doubling of the number of responses. We find that reasoning evaluators achieve better performance compared to their direct counterparts given a fixed test-time compute budget: see \autoref{sec:appendix-approx-compute} for details of our FLOPs calculations. \textbf{(right)} Combining outcome and process scores yields further gains.
    }
    \label{fig:bon2}
\end{figure*}

\begin{table*}[!t]
\centering
\resizebox{0.85\textwidth}{!}{
\begin{tabular}{lccccccc}
\toprule
\textbf{Model} & \textbf{$N=1$} & \textbf{$N=2$} & \textbf{$N=4$} & \textbf{$N=8$} & \textbf{$N=16$} & \textbf{$N=32$}& \textbf{$N=64$}\\
\midrule
\multicolumn{1}{l}{\textbf{Direct Outcome Evaluators (ORMs)}} \\
\midrule
Skywork-Reward-Llama-3.1-8B-v0.2 & 38.2 & 41.4 & 43.4 & 44.6 & 45.5 & 45.2 & 44.8 \\
Skywork-Reward-Gemma-2-27B-v0.2 & 38.2 & 41.8 & 43.4 & 44.8 & 45.5 & 45.6 & 45.4 \\
\midrule
\multicolumn{1}{l}{\textbf{Direct Process Evaluators (PRMs)}} \\
\midrule
math-shepherd-mistral-7b-prm & 38.2 & 41.6 & 42.7 & 43.1 & 43.7 & 43.5 & 43.2 \\
Skywork-o1-Open-PRM-Qwen-2.5-1.5B & 38.2 & 42.8 & 44.2 & 45.9 & 46.5 & 46.5 & 46.7 \\
Skywork-o1-Open-PRM-Qwen-2.5-7B & 38.2 & 42.9 & 45.3 & 47.5 & 48.4 & 48.7 & 49.9 \\
RLHFlow/Llama3.1-8B-PRM-Mistral & 38.2 & 40.2 & 40.0 & 39.4 & 38.4 & 37.2 & 35.5 \\
RLHFlow/Llama3.1-8B-PRM-Deepseek & 38.2 & 40.5 & 40.4 & 40.4 & 40.0 & 38.8 & 37.8 \\
Qwen2.5-Math-7B-PRM800K & 38.2 & 41.6 & 43.4 & 45.1 & 45.2 & 45.0 & 44.6 \\
Qwen2.5-Math-PRM-7B & 38.2 & 42.3 & 44.0 & 45.8 & 47.8 & 48.2 & 48.7 \\
Qwen2.5-Math-PRM-72B & 38.2 & 42.9 & 45.4 & 48.2 & 48.9 & 49.8 & 50.6 \\
\midrule
\multicolumn{1}{l}{\textbf{Non-Reasoning Generative Evaluators}} \\
\midrule
Llama3-8B-CLoud-RM & 38.2 & 41.8 & 42.7 & 43.8 & 43.7 & 43.1 & 42.5 \\
prometheus-7b-v2.0 & 38.2 & 40.4 & 41.3 & 41.7 & 41.5 & 40.7 & 39.7 \\
prometheus-8x7b-v2.0 & 38.2 & 40.3 & 40.8 & 41.9 & 41.6 & 41.2 & 40.8 \\
Qwen2.5-72B-Instruct & 38.2 & 42.3 & 44.1 & 45.9 & 45.2 & 45.8 & 45.6 \\
\midrule
\multicolumn{1}{l}{\textbf{Reasoning Outcome Evaluators}} \\
\midrule
\rowcolor{lightpink} DeepSeek-R1-Distill-Qwen-7B & 38.2 & 42.7 & 46.6 & 48.7 & - & - & - \\
\rowcolor{lightpink} DeepSeek-R1-Distill-Qwen-32B & 38.2 & \underline{43.9} & \underline{47.7} & \underline{51.1} & - & - & - \\
\midrule
\multicolumn{1}{l}{\textbf{Single-step Reasoning Process Evaluators}} \\
\midrule
\rowcolor{lightgray} DeepSeek-R1-Distill-Qwen-7B & 38.2 & 41.3 & 44.2 & 45.6 & - & - & - \\
\rowcolor{lightgray} DeepSeek-R1-Distill-Qwen-32B & 38.2 & 41.7 & 44.9 & 48.1 & - & - & - \\
\midrule
\multicolumn{1}{l}{\textbf{Reasoning Process Evaluators}} \\
\midrule
\rowcolor{lightblue} DeepSeek-R1-Distill-Qwen-7B & 38.2 & 42.6 & 45.5 & 48.3 & - & - & - \\
\rowcolor{lightblue} DeepSeek-R1-Distill-Qwen-32B & 38.2 & 43.6 & 46.9 & 50.3 & - & - & - \\
\midrule
\multicolumn{1}{l}{\textbf{Reasoning Process + Outcome Evaluators}} \\
\midrule
\rowcolor{darkgray} DeepSeek-R1-Distill-Qwen-7B & 38.2 & 43.3 & 46.6 & 50.1 & - & - & - \\
\rowcolor{darkgray} DeepSeek-R1-Distill-Qwen-32B & 38.2 & \textbf{44.4} & \textbf{48.5} & \textbf{52.0} & - & - & - \\
\bottomrule
\end{tabular}
}
\caption{Full results for Best-of-8 (reasoning evaluators) and Best-of-64 (all other evaluators) experiments using direct outcome evaluators, direct process evaluators, non-reasoning generative evaluators, \thinpinkcolorbox{lightpink}{reasoning outcome evaluators}, \thinlightgraycolorbox{lightgray}{single-step reasoning process evaluators}, \thinbluecolorbox{lightblue}{reasoning process evaluators}, and \thindarkgraycolorbox{darkgray}{reasoning outcome + process evaluators}. Results were obtained by averaging scores across all 7 evaluation benchmarks and 3 generators as described in \autoref{section:problem_solving}.}
\label{tab:bon_extended1}
\end{table*}

\begin{table*}[!t]
\centering
\resizebox{0.6\textwidth}{!}{
\begin{tabular}{lcccc}
\toprule
\textbf{Model} & \textbf{$N=1$} & \textbf{$N=2$} & \textbf{$N=4$} & \textbf{$N=8$}\\
\midrule
\multicolumn{5}{c}{\textbf{Splitting}: Heuristic-based, \textbf{Aggregation}: min} \\
\midrule
Qwen2.5-Math-PRM-7B &  38.2 & 42.3 & 44.0 & 45.8 \\
Qwen2.5-Math-PRM-72B &  38.2 & 42.9 & 45.4 & 48.2\\
\midrule
\multicolumn{5}{c}{\textbf{Splitting}: Heuristic-based, \textbf{Aggregation}: mean\_logits} \\
\midrule
Qwen2.5-Math-PRM-7B &  38.2 & 40.7 & 41.4 & 41.7 \\
Qwen2.5-Math-PRM-72B &  38.2 & 41.9 & 43.3 & 43.8\\
\midrule
\multicolumn{5}{c}{\textbf{Splitting}: Model-based, \textbf{Aggregation}: min} \\
\midrule
Qwen2.5-Math-PRM-7B &  38.2 & 42.6 & 45.00 & 47.1 \\
Qwen2.5-Math-PRM-72B &  38.2 & 43.2 & 45.8 & 49.9\\
\bottomrule
\end{tabular}
}
\caption{Ablation results of applying different splitting and aggregation functions to direct process evaluators.}
\label{tab:bon_extended2}
\end{table*}

\section{Extended: Experimental Settings}\label{appendix:experiment_setting}

\subsection{Model List for ProcessBench Experiments (\autoref{section:scaling})}\label{appendix:model_list_scaling}

We examine three varieties of models in our ProcessBench experiments:
\begin{itemize}[leftmargin=*]
\item \textbf{Direct PRMs:} We experiment with 10 different direct PRMs, representing the state-of-the-art on ProcessBench, from families including \texttt{math-shepherd-mistral}~\citep{wang2024math}, \texttt{Skywork}~\citep{skyworkopeno12024}, \texttt{RLHFlow}~\citep{xiong2024rlhflowmath}, \texttt{EurusPRM}~\citep{cui2025process}, and \texttt{Qwen2.5-Math-PRM}~\citep{zhang2025lessons}.
\item \textbf{Instruction-tuned Models:} These are models that have been trained using supervised fine tuning and/or RLHF, but have not been explicitly trained for reasoning.
We experiment with models from the \texttt{Llama-3.1}~\citep{dubey2024llama3herdmodels}, \texttt{Llama-3.3}~\citep{dubey2024llama3herdmodels}, and \texttt{Qwen2.5}~\citep{qwen2.5} families. We also experiment with \texttt{GPT-4-0806} as an outcome evaluator.
\item \textbf{Reasoning Models:} These are models that have been explicitly trained to perform reasoning using RL, or distilled from models trained to perform reasoning.
We examine models from the \texttt{DeepSeek-R1-Distill-Qwen}~\citep{guo2025deepseek} and \texttt{QwQ}~\citep{qwq-32b-preview} families. We also experiment with \texttt{o1-mini} as a reasoning evaluator.
\end{itemize}

We report results from a subset of these models in the main text: see \autoref{tab:processbench}. We include results from all listed models in \autoref{tab:processbench_extended}.

Note that we do not experiment with fine-tuned generative evaluators such as Prometheus~\citep{kim2023prometheus, kim2024prometheus}, CLoud-RM~\citep{ankner2024critique} and GenRM~\citep{zhang2024generative} as these are trained outcome evaluators that cannot be readily employed to detect process errors, as is required by ProcessBench.

\subsection{Model List for Best-of-$N$ Experiments (\autoref{section:problem_solving})}
\label{appendix:model_list_bon}

\begin{itemize}[leftmargin=*]
    \item \textbf{Direct outcome evaluators}: We experiment with \texttt{Skywork-Reward-Llama-3.1-8B-v0.2}~\citep{liu2024skywork} and \texttt{Skywork-Reward-Gemma-2-27B-v0.2}~\citep{liu2024skywork}, which are the state-of-the-art direct outcome evaluators on RewardBench~\citep{lambert2024rewardbench} (a widely used benchmark to assess ORMs).
    \item \textbf{Direct process evaluators}: We experiment with \texttt{math-shepherd-mistral}~\citep{wang2024math}, \texttt{Skywork}~\citep{skyworkopeno12024}, \texttt{RLHFlow}~\citep{dong2024rlhf}, \texttt{EurusPRM}~\citep{cui2025process}, and \texttt{Qwen2.5-Math-PRM}~\citep{zhang2025lessons}.
    \item \textbf{Fine-tuned generative evaluators}: While, like reasoning models, these models also produce CoT they, unlike reasoning models, produce short CoT that lack complex reasoning patterns (e.g. self-verification, self-correction, backtracking) as they are not trained to reason using RL. We experiment with \texttt{Llama3-8B-CLoud-RM}~\citep{ankner2024critique} and the \texttt{Prometheus 2} family~\citep{kim2024prometheus2}, which are generative models trained specifically for outcome evaluation, as well as \texttt{Qwen2.5-72B-Instruct}~\citep{qwen2.5}, which we prompt to act as a generative evaluator. We use these models as reasoning outcome evaluators.
    \item \textbf{Reasoning outcome evaluators}: We prompt reasoning models \texttt{DeepSeek-R1-Distill-\\Qwen-7B} and \texttt{DeepSeek-R1-Distill-Qwen-32B} to act as reasoning outcome evaluators.
    \item \textbf{Reasoning process evaluator}: We prompt reasoning models \texttt{DeepSeek-R1-Distill-\\Qwen-7B} and  \texttt{DeepSeek-R1-Distill-Qwen-32B} to act as reasoning process evaluators.
    \item \textbf{Reasoning process + outcome evaluator}: We experiment with the reasoning models \texttt{DeepSeek-R1-Distill-Qwen-7B} and \texttt{DeepSeek-R1-Distill-Qwen-32B}, using identical models for both the outcome and process evaluation components in each case.
\end{itemize}

\section{Additional Results for ProcessBench Experiments (\autoref{section:scaling})}\label{appendix:scaling_extended}

We include the full experimental results for our ProcessBench experiments (\autoref{section:scaling}) in \autoref{tab:processbench_extended}.

\begin{table*}[!t]
\centering
\resizebox{\textwidth}{!}{
\begin{tabular}{lcccccc}
\toprule
\textbf{Model} & \textbf{GSM8K} & \textbf{MATH} & \textbf{Olym. Bench} & \textbf{Omni-MATH} & \textbf{Avg. F1} & \textbf{$\Delta$}\\
\midrule
\multicolumn{7}{c}{\textbf{Direct Process Evaluator (PRMs)}} \\
\midrule
math-shepherd-mistral-7b-prm &   47.9  & 29.5  & 24.8  & 23.8 & 31.5& -\\
Skywork-o1-Open-PRM-Qwen-2.5-1.5B &  57.9 & 48.0  & 16.5 & 18.9 & 35.3 & -\\
Skywork-o1-Open-PRM-Qwen-2.5-7B &  70.8  & 53.6  & 22.9 & 21.0 & 42.1 & -\\
RLHFlow/Llama3.1-8B-PRM-Mistral &  50.4 & 33.4 & 13.8  & 15.8 & 28.4 & -\\
RLHFlow/Llama3.1-8B-PRM-Deepseek &  38.8& 33.8 & 16.9  & 16.9 & 26.6 & -\\
EurusPRM-Stage1 (7B) &  44.3		& 35.6 & 21.7		& 23.1	& 31.2 & -\\
EurusPRM-Stage2 (7B) &  47.3		& 35.7	& 21.2		& 20.9	& 31.3 & -\\
Qwen2.5-Math-7B-PRM800K & 68.4 & 62.5 & 50.4 & 43.6 & 56.2 & - \\
Qwen2.5-Math-PRM-7B &   82.4 	& 77.6		& 67.5	& 66.3	& 73.5 & -\\
Qwen2.5-Math-PRM-72B &  \textbf{87.3}	& \textbf{80.6}& \textbf{74.3}		& \textbf{71.1}	& \textbf{78.3} & -\\
\midrule
\multicolumn{7}{c}{\textbf{Single-step Reasoning Process Evaluator}} \\
\midrule
\noalign{\vspace{4pt}}
\multicolumn{7}{l}{\textbf{Instruction-tuned Models}} \\
Llama-3.1-8B-Instruct   & 24.0& 15.5 & 9.7  & 10.1 & 14.8 & 0.0\\
Llama-3.3-70B-Instruct$^{\dagger}$   & \textbf{82.9} & 59.4  & 46.7  & 43.0 & 58.0 & -\\
Qwen2.5-32B-Instruct & 63.8 & 47.5 & 35.9 & 32.7 & 45.0 & 0.0\\
Qwen2.5-Math-72B-Instruct$^{\dagger}$  & 65.8 & 52.1  & 32.5 & 31.7 & 45.5 & -\\
Qwen2.5-72B-Instruct$^{\dagger}$  & 76.2 & \textbf{61.8}& \textbf{54.6} & \textbf{52.2} & \textbf{61.2} & -\\
\noalign{\vspace{4pt}}
\multicolumn{7}{l}{\textbf{Reasoning Models}} \\
\rowcolor{lightgray} DeepSeek-R1-Distill-Qwen-7B  & 68.3 & 61.1 & 48.3 & 40.4 &54.5 & 0.0\\
\rowcolor{lightgray} DeepSeek-R1-Distill-Qwen-32B  & \textbf{83.9} & \textbf{78.1} & \textbf{72.4} & 67.7 & \textbf{75.5} & 0.0\\
\rowcolor{lightgray} QwQ-32B-Preview   & 77.5 & 58.9 & 31.2  & 35.8 & 50.9 & 0.0\\
\rowcolor{lightgray} QwQ-32B   & 79.5 & 77.5 & 71.5  & \textbf{69.4} & 74.5 & 0.0\\
\noalign{\vspace{4pt}}
\multicolumn{7}{l}{\textbf{Reasoning Models} (Self-Consistency)} \\
\rowcolor{lightgray} DeepSeek-R1-Distill-Qwen-7B & 69.3&67.9&54.8 & 51.5 & 60.9 & +6.4\\
\rowcolor{lightgray} DeepSeek-R1-Distill-Qwen-32B & 82.2 & \textbf{80.4} & \textbf{76.2} & 72.5 & \textbf{77.8} & +2.3\\
\rowcolor{lightgray} QwQ-32B-Preview & \textbf{88.0} & 78.7 & 57.8& 61.3 & 71.5 & +20.6\\
\rowcolor{lightgray} QwQ-32B & 81.0 & 78.8 & 74.4 & \textbf{72.8} & 76.8 & +2.3\\
\noalign{\vspace{4pt}}
GPT-4-0806$^{\dagger}$ & 79.2& 63.6 & 51.4 & 53.5 & 61.9 &-\\
o1-mini$^{\dagger}$ & \textbf{93.2} & \textbf{88.9} & \textbf{87.2} & \textbf{82.4} & \textbf{87.9} &-\\
\midrule
\multicolumn{7}{c}{\textbf{Reasoning Process Evaluator (Ours)}} \\
\midrule
\noalign{\vspace{4pt}}
\multicolumn{7}{l}{\textbf{Instruction-tuned Models} (Multi-step Process Evaluation)} \\
Llama-3.1-8B-Instruct & 35.2& 22.8  & 12.6  & 17.9 & 22.1&+7.3\\
Qwen2.5-32B-Instruct& \textbf{70.1} & \textbf{61.7} & \textbf{54.2} & \textbf{53.9} & \textbf{60.0} & +15.0\\
\noalign{\vspace{4pt}}
\multicolumn{7}{l}{\textbf{Reasoning Models} (Multi-step Process Evaluation)} \\
\rowcolor{lightblue} DeepSeek-R1-Distill-Qwen-7B & 75.5 & 67.3 & 59.8 & 56.6 & 64.8&+10.3\\
\rowcolor{lightblue} DeepSeek-R1-Distill-Qwen-32B &  80.3 & 82.2 & \textbf{77.0} & 75.0 & 78.6 & +3.1\\
\rowcolor{lightblue} QwQ-32B-Preview  & \textbf{81.7} & 79.3 & 70.3 & 69.8 & 75.3 & +24.4\\
\rowcolor{lightblue} QwQ-32B  & 81.5 & \textbf{83.6} & 76.8 & \textbf{75.1} & \textbf{79.3} & +4.8\\
\noalign{\vspace{4pt}}
\multicolumn{7}{l}{\textbf{Reasoning Models} (Multi-step Process Evaluation + Self-Consistency)} \\
\rowcolor{lightblue} DeepSeek-R1-Distill-Qwen-7B & 80.1& 75.1 & 69.3 & 70.4 & 73.7 & +19.2\\
\rowcolor{lightblue} DeepSeek-R1-Distill-Qwen-32B & 86.6 & 85.4 & 78.9 & \textbf{80.3} & \textbf{82.8} & +7.3\\
\rowcolor{lightblue} QwQ-32B-Preview &  \textbf{86.8} & \textbf{85.7}  & 79.0 & 78.0 & 82.4&+31.5\\
\rowcolor{lightblue} QwQ-32B &  85.0 & 85.6 & \textbf{79.4} & 78.0 & 82.0 & +7.5\\
\bottomrule
\end{tabular}
}
\caption{Full experimental results for our ProcessBench experiments (\autoref{section:scaling}).}
\vspace{-3mm}
\label{tab:processbench_extended}
\end{table*}

\section{Additional Results and Discussion for Best-of-$N$ Experiments (\autoref{section:problem_solving})}\label{appendix:bon_extended}

We include the full experimental results for our Best-of-$N$ experiments (\autoref{section:problem_solving}) in \autoref{tab:bon_extended1}.

\subsection{Findings from Experiments}

\textbf{Evaluation-time Scaling is effective with smaller-sized evaluators as well}. \quad Similar to \autoref{fig:bon1} with larger-sized evaluators, \autoref{fig:bon2} shows the results of employing smaller-sized evaluators in the Best-of-$N$ setting. The findings from \autoref{section:problem_solving} maintain the same: (1) reasoning evaluators (Best-of-8) outperform or match their direct evaluator counter parts (Best-of-64) while using less amount of compute, (2) reasoning process + outcome evaluation can boost performance, and (3) reasoning evaluators are especially effective for coding.

\textbf{Multi-step process evaluation outperforms single-step process evaluation}. \quad Next, as shown in Table~\ref{tab:bon_extended1}, we compare reasoning process evaluators with single-step reasoning process evaluators (see \autoref{subsec:prm} for a detailed explanation of the difference). Results show that even when employing the same LM as the evaluator, evaluating each step individually is superior to evaluating all the steps at once, supporting the strength of our approach and effectiveness of evaluation-time scaling.

\textbf{Reasoning outcome evaluators outperform specially-trained outcome evaluators.}. \quad Then, as shown in Table~\ref{tab:bon_extended1}, we compare the effectiveness of employing reasoning models as outcome evaluators over using specially-trained outcome evaluators such as CLoud-RM~\citep{ankner2024critique} and Prometheus 2~\citep{kim2024prometheus}. Results show that reasoning models are very effective in our Best-of-$N$ setting. This is notable because it hints that employing LMs with stronger problem-solving capabilities as evaluators is more important than inducing evaluation capabilities through training. Future work could explore recipes for training reasoning models as evaluators.

\textbf{Model-based splitting is effective for direct process evaluators as well}. \quad Lastly, we ablate the effect of applying model-based splitting and the \texttt{mean\_logits} aggregation function to direct process evaluators. Note that model-based splitting requires the usage of an LM ($M_{split}$) to segment the response into steps, it requires additional compute. Results in \autoref{tab:bon_extended2} show that (1) applying a model-based splitting approach is effective and (2) using the mean\_logits aggregation function is not.

\begin{table*}[ht]
\centering
\resizebox{0.4\textwidth}{!}{
\begin{tabular}{cccc}
\toprule
\textbf{N} & \textbf{Mean Score} & \textbf{Std Dev} & \textbf{96\% CI}\\
\midrule
1 & 38.14 & 1.712 & (34.31, 41.96) \\
2 & 43.98 & 1.752 & (40.06, 47.90) \\
4 & 47.90 & 1.783 & (43.92, 51.89) \\
8 & 51.58 & 1.782 & (47.60, 55.57) \\
\bottomrule
\end{tabular}}
\caption{Statistical analysis of mean\_logit scores for different sample sizes in Best-of-N evaluation}
\label{tab:stats}
\end{table*}

\subsection{Checking statistical significance of results}\label{appendix:statistical_test}

We run the Best-of-8 experiment twice, with a reasoning evaluator (DeepSeek-R1-Distill-Qwen-32B) calculating scores on 8 responses per instance. For N=8, there are 2 scores; for N=4, there are 4 scores (index 0$\sim$3 from first run, index 4$\sim$7 from first run, index 0$\sim$3 from second run, index 4$\sim$7 from second run); and so on. As shown in \autoref{tab:stats}, all scores across N=1,2,4,8 are statistically significant at the 96\% confidence level.

\begin{table*}[!t]
\centering
\footnotesize
\resizebox{\textwidth}{!}{
\begin{tabular}{l c ccc ccc c}
\toprule
\textbf{Evaluator} & \textbf{\# Generator Responses} & \multicolumn{3}{c}{\textbf{Llama-3.1-70B-Instruct}} & \multicolumn{3}{c}{\textbf{DeepSeek-R1-Distill-Qwen-7B}} & \textbf{Total Avg} \\
\cmidrule(lr){3-5} \cmidrule(lr){6-8}
& & Economics & Engineering & Law & Economics & Engineering & Law & \\
\midrule
\multicolumn{9}{c}{\textbf{Baselines}} \\
\midrule
Greedy N=1 & 1 & 74.5 & 47.0 & 45.5 & 69.5 & 53.0 & 17.5 & 51.17 \\
Skywork-8B (ORM) & 64 & 78.0 & 47.0 & 50.0 & 76.5 & 55.5 & 23.5 & 55.08 \\
Skywork-27B (ORM) & 64 & 80.5 & 42.5 & 51.0 & 74.5 & \textbf{58.0} & 23.5 & 54.92 \\
Qwen-PRM-7B & 64 & 76.0 & 50.5 & 47.5 & 74.0 & 55.0 & 21.5 & 54.08 \\
Qwen-PRM-72B & 64 & 74.5 & 41.5 & \textbf{48.0} & 78.0 & 52.0 & 26.0 & 53.33 \\
\midrule
\multicolumn{9}{c}{\textbf{Ours}} \\
\midrule
DeepSeek-R1-Distill-Qwen-7B (Process + Outcome) & 8 & 79.0 & 51.5 & 47.0 & 75.0 & 52.0 & 28.5 & 55.33 \\
DeepSeek-R1-Distill-Qwen-32B (Process + Outcome) & 8 & \textbf{81.0} & \textbf{56.5} & 47.5 & \textbf{79.0} & 54.0 & \textbf{32.0} & \textbf{58.33} \\
\bottomrule
\end{tabular}}
\caption{\textbf{Reasoning process evaluators are effective outside of math and code as well:} Performance across Economics, Engineering, and Law subsets from MMLU-Pro~\citep{wang2024mmlu} in Best-of-$N$ experiment setting using different evaluators. Our approach (DeepSeek-R1-Distill-Qwen-7B/32B with Process + Outcome evaluation) outperforms other evaluators while using less number of responses from generators.}
\label{tab:outside_math_code}
\end{table*}

\subsection{Testing on domains outside of math and code}\label{appendix:outside_of_math_and_code}

\textbf{Experimental setting} To evaluate whether our reasoning process evaluator extends beyond math and coding tasks, we conduct experiments on 500 samples from each of the economics, engineering, and law subsets of MMLU-Pro~\citep{wang2024mmlu}. To sample responses, we use both a reasoning model (DeepSeek-R1-Distill-Qwen-7B) and a non-reasoning model (Llama-3.1-70B-Instruct). For each question, we sample 64 candidate responses from both generators.

\textbf{Experimental results} The results are shown in \autoref{tab:outside_math_code}. While baseline evaluators (ORMs and PRMs) operate on all 64 candidates, our reasoning-based evaluator processes only 8 responses. Despite this smaller candidate set, the results show that our 7B evaluator outperforms both the 8B ORM and the 7B PRM, and our 32B evaluator outperforms both the 27B ORM and the 32B PRM. These findings, consistent with the trends observed in \autoref{fig:bon1} and \autoref{fig:bon2}, demonstrate that evaluation-time scaling with reasoning evaluators remains effective across diverse domains, including economics, engineering, and law, thereby confirming that our methodology generalizes beyond math and coding.

\begin{table*}[!t]
\centering
\footnotesize
\resizebox{\textwidth}{!}{
\begin{tabular}{p{0.17\textwidth} p{0.29\textwidth} p{0.22\textwidth} p{0.30\textwidth}}
\toprule
\textbf{Paper} &
\textbf{Method for Scaling Test-time Compute on Generator side} &
\textbf{Method for Scaling Test-time Compute on Evaluator side} &
\textbf{Investigated Trade-off} \\
\midrule
\textbf{\citet{snell2024scaling}} &
\textbf{(A)} Generating Multiple Responses (Parallel), \textbf{(B)} Beam/Lookahead search using a PRM (Parallel), \textbf{(C)} Sequential Revision (Sequential) &
\textbf{(X)} Using a PRM &
\textbf{``A{+}X'' vs ``B{+}X''}: Test-time scaling for search with verifiers;\quad
\textbf{``A{+}X'' vs ``C{+}X''}: Test-time scaling with revisions. \\
\midrule
\textbf{\citet{muennighoff2025s1}} &
\textbf{(A)} Generating Multiple Responses (Parallel), \textbf{(B)} Budget Forcing (Sequential) &
\textbf{(X)} Using a PRM &
\textbf{``A{+}X'' vs ``B{+}X''}: Parallel vs sequential scaling. \\
\midrule
\textbf{\citet{brown2024large}} &
\textbf{(A)} Generating Multiple Responses &
\textbf{(X)} Using an ORM &
\textbf{``A{+}X''}: Scaling laws as the number of responses increases. \\
\midrule
\textbf{Ours} &
\textbf{(A)} Generating Multiple Responses (Parallel) &
\textbf{(X)} Using an ORM or PRM;\; \textbf{(Y)} Applying self-consistency to the evaluator (Parallel);\; \textbf{(Z)} Utilizing a reasoning model as a process evaluator (Sequential) &
\textbf{``X'' vs ``Y'' vs ``Z''}: Evaluation-time scaling (Sec.~3);\quad
\textbf{``A{+}X'' vs ``A{+}Z''}: Translating improved reasoning to problem solving (Sec.~4). \\
\bottomrule
\end{tabular}}
\caption{\textbf{Test-time scaling on generator vs.\ evaluator:} Summary of methods and trade-offs across prior work and ours. Unlike prior work that mainly scales the generator while fixing the evaluator, we are the first to explicitly study scaling \emph{evaluator} test-time compute. We show that scaling evaluation-time strategies can directly improve evaluator performance, opening new directions for test-time evaluation research. Generator-side variants include (A) multiple responses (parallel), (B) beam/lookahead with a PRM (parallel), (C) sequential revision. Evaluator-side variants include (X) a PRM, (Y) self-consistency on the evaluator (parallel), and (Z) a reasoning model as a process evaluator (sequential).}
\label{tab:literature_search}
\end{table*}

\section{Related work}
\label{sec:related_work}


\subsection{Scaling Test-time Compute}

Increasing compute by enlarging model size or expanding training data has long been one of the key methods to improve LM performance during training time~\citep{kaplan2020scaling,hoffmann2022empirical,longpre2023flan,kim2023cot}. However, as it becomes increasingly difficult to obtain high-quality data sufficient for steady advancement of LM performance, a new paradigm has gained attention: scaling compute at test-time instead of training-time~\citep{villalobos2024position,welleck2024from,wu2024scaling}. This approach is attracting interest as a method that can enhance LM performance in a way that complements training time compute. The main approaches to scaling up test-time compute include, first, leveraging sufficient compute at test-time by training reasoning models that generate longer and qualitatively different Chain-of-Thought (CoT) compared to existing chat models~\citep{guo2025deepseek,yeo2025demystifying,muennighoff2025s1}, and second, using inference-time algorithms such as Best-of-$N$ sampling at test-time~\citep{sun2024easytohard,welleck2024from}. Existing works on test-time compute have primarily focused on improving LM's problem-solving capability, whereas we focus on scaling compute for evaluation to enhance evaluators' capabilities by assessing each response step with process evaluation and generating long CoT for precise evaluation. 

\autoref{tab:literature_search} summarizes the difference between representation works on test-time scaling (on the generator side) versus our approach on scaling evaluation-time compute.

\subsection{Language Model Evaluators}

Accurately verifying the outputs generated by a language model (LM) is crucial for understanding the types of errors it frequently makes and identifying its limitations~\citep{liang2023holistic,mondorf2024beyond,zheng2024processbench, lee2025evaluating}.
Recently, evaluators—LMs that assess the quality of a given response (also referred to as verifiers, reward models, or judges in the literature)—have gained significant attention for their ability to provide precise assessments of LM outputs~\citep{zheng2024judging,ye2024flask,kim2023prometheus,lambert2024rewardbench,gu2024survey}.
Evaluators are not only used for benchmarking purposes but also for enhancing the LM's problem solving capabilities~\citep{cobbe2021training,uesato2022solving,lightman2024let,wang2024math,sun2024easytohard,wu2024better}.

When an evaluator fails to assess accurately, it may result in unintended consequences for the purpose it is serving~\citep{gao2023scaling,coste2024reward,moskovitz2024confronting}.
For example, if an evaluator fails to provide accurate judgments, even if a specific LM being evaluated performs well, its true capabilities may be misrepresented due to the errors stemming from the evaluator's limitations~\citep{dubois2024length,arenahard2024,kim2024biggen}. Also, when integrating an evaluator into an inference-time algorithm, the imperfection of the evaluator might result in diminishing returns even when using more test-time compute~\citep{gao2023scaling,rafailov2024scaling}. These limitations highlight the need for more robust evaluators that can generalize in diverse contexts. While \citet{kalra2025verdict} has examined debate-based strategies and usage of larger models as evaluators to scale up evaluation-time compute, our work specifically focuses on `using reasoning models as process evaluators' to demonstrate the effectiveness of evaluation-time scaling.

\subsection{Future Directions}

Looking ahead, we envision our research enabling advances in two areas. First, evaluation-time scaling can provide better training signals. In particular, it is widely known that generators often develop undesirable traits through reward model over-optimization when given imprecise rewards during reinforcement learning~\citep{stiennon2020learning,bai2022training,ouyang2022training,huang202237}; investigating whether reasoning process evaluators can mitigate this represents a promising direction. Second, future work could explore whether reasoning evaluators can be improved through training. Existing trained evaluators do not leverage the long CoTs that have proven effective in this work, yet we believe that training such models may be key to further enhancing LM evaluation capabilities.

\section{Approximation of Test-Time Compute}
\label{sec:appendix-approx-compute}

For approximating inference-time compute as in \autoref{fig:bon1} and \autoref{fig:bon2}, we follow \citet{snell2024scaling} and \citet{son2025linguistic}. Specifically, inference compute cost can be asymptotically approximated by:
\begin{equation}
C \in O(N \times L),
\end{equation}
where $C$ is the computation cost, $N$ is the number of parameters and $L$ is the number of tokens. Therefore, we use $N \times L$ as a relative inference compute for a single inference call. 

For instance, consider a Best-of-8 case where the generator of size 70B generates total 1,000 tokens in average (\textbf{generation-time compute} for response), and the reasoning outcome evaluator of size 7B generates total 3,000 tokens in average (\textbf{evaluation-time compute} for CoT and judgment). In this case, the approximate inference-time compute can be calculated as:
\begin{equation}
\begin{split}
&8 \times ((70 \times 10^9 \times 1000) + (7 \times 10^9 \times 3000)) \\
&\qquad = 7.28 \times 10^{17}
\end{split}
\end{equation}

On a high level, when we break down inference-time compute into generator-time compute and evaluation-time compute, $70 \times {10^9}\times 1000$ corresponds to the generation-time compute and $ 7\times {10^9} \times 3000$ corresponds to the evaluation-time compute. Therefore, Best-of-8 with reasoning process evaluators (that spends more evaluation-time compute than generation-time compute) requires similar inference-time compute compared to Best-of-64 with direct evaluators (that spends more generation-time compute than evaluation-time compute).

\section{Formal explanation of why evaluation-time scaling can improve the generator's performance}\label{appendix:formal_explanation}

We provide a more formal explanation for why evaluation-time scaling can be more effective than the conventional approach of sampling a larger number of responses with a weaker evaluator. The key intuition is that a stronger evaluator mitigates the phenomenon of reward model over-optimization, where imperfect evaluators overvalue certain responses due to noise or bias.  

Let $u(x)$ denote the oracle quality of a candidate response $x$. An evaluator provides a surrogate score  

\begin{equation}
E(x) = u(x) + \delta(x),
\end{equation}

where $\delta(x)$ captures evaluation error (biases or noise). In Best-of-$N$, given $N$ candidates $Y = \{y_1, \dots, y_N\}$, the oracle-best candidate is  

\begin{equation}
y^*_N = \arg\max_{y_i \in Y} u(y_i),
\end{equation}

while the candidate chosen by the evaluator is  

\begin{equation}
\arg\max_{y_i \in Y} \big(u(y_i) + \delta(y_i)\big).
\end{equation}

Because the evaluator is imperfect, the selected response may be suboptimal if it receives an erroneously high $\delta(y_i)$. The probability of such mis-selection grows with $N$, since larger candidate sets increase the chance of some $y_i$ having a large positive error. This effect, often referred to as \emph{reward model over-optimization} or \emph{reward hacking}, undermines the benefits of scaling the generator alone.  

Suppose we instead use an improved evaluator  

\begin{equation}
E'(x) = u(x) + \delta'(x),
\end{equation}

where $\delta'(x)$ has lower variance or higher fidelity with respect to $u(x)$. In this case, the probability of correctly selecting the oracle-best candidate improves:  

\begin{equation}
\begin{split}
&P\!\left(\arg\max_{i} E'(y_i) = y^*_N\right) \;> \\ 
&\qquad P\!\left(\arg\max_{i} E(y_i) = y^*_N\right).
\end{split}
\end{equation}

Crucially, our results show that even with fewer candidates ($n \ll N$), scaling evaluation-time compute with a stronger evaluator can yield  

\begin{equation}
\begin{split}
&P\!\left(\arg\max_{i} E'(y_i) = y^*_n\right) \;> \\ 
&\qquad P\!\left(\arg\max_{i} E(y_i) = y^*_N\right).
\end{split}
\end{equation}

Thus, despite reducing the number of samples, improved evaluation quality can more than compensate for the loss in oracle score.  

This explanation aligns with our experimental results provided in \autoref{section:problem_solving}. First, in the low-budget regime (e.g., 1–2 candidates), investing in evaluation compute provides little benefit, since oracle quality is too low for the evaluator to meaningfully distinguish among responses. In this regime, additional sampling dominates. Second, as the number of candidates grows (16–64), a weaker evaluator (e.g., PRM) becomes increasingly vulnerable to mis-selection, while our approach continues to benefit from reduced error. Finally, scaling evaluation-time compute produces smooth gains in general: when moving from 1 to 8 responses, both the oracle quality improves and the stronger evaluator reliably identifies the best candidate, leading to consistent upward scaling curves.  

\section{Self-Evaluation of Reasoning Models}\label{appendix:self_eval}

\begin{table*}[!t]
\centering
\footnotesize
\resizebox{\textwidth}{!}{
\begin{tabular}{lcccc}
\toprule
\textbf{Generator} & \textbf{$N = 1$} & \textbf{$N = 8$} & \textbf{Oracle} & \textbf{Gap Recovered (\%)} \\
\midrule
\multicolumn{5}{c}{\textbf{AIME24}} \\
\midrule
Eurus-2-SFT & 13.3 & 20.0 & 20.0 & \textbf{100.0} \\
Llama3.1-70B-Instruct & 16.7 & 23.3 & 36.7 & 33.0 \\
Qwen2.5-7B-Instruct & 10.0 & 16.7 & 23.3 & 50.4 \\
\midrule
Self-Evaluation (Outcome Eval on CoT + Process Eval on summary) & 50.0 & 73.3 & 83.3 & \underline{70.0} \\
Self-Evaluation (Outcome Eval on summary + Process Eval on summary) & 50.0 & 66.7 & 83.3 & 50.2 \\
\midrule
\multicolumn{5}{c}{\textbf{AMC23}} \\
\midrule
Eurus-2-SFT & 31.1 & 45.3 & 62.7 & 44.9 \\
Llama3.1-70B-Instruct & 26.8 & 45.3 & 65.1 & \underline{48.3} \\
Qwen2.5-7B-Instruct & 36.6 & 51.6 & 69.9 & 45.0 \\
\midrule
Self-Evaluation (Outcome Eval on CoT + Process Eval on summary) & 85.5 & 88.0 & 92.8 & 34.2 \\
Self-Evaluation (Outcome Eval on summary + Process Eval on summary) & 85.5 & 89.2 & 92.8 & \textbf{50.7} \\
\bottomrule
\end{tabular}}
\caption{\textbf{Reasoning models can self-evaluate its response effectively when functioning as a reasoning process evaluator:} We prompt DeepSeek-R1-Distill-Qwen-7B to act as both a generator and its own Best-of-$N$ reasoning process + outcome evaluator (\textbf{self-evaluation}). We also report the improvements when using Eurus-2-SFT, Llama3.1-70B-Instruct, and Qwen2.5-7B-Instruct as a generator and DeepSeek-R1-Distill-Qwen-7B as an evaluator for relative comparison. We measure performance improvements (between Best-of-1 and Best-of-8) as a percentage of the gap between Best-of-1 and oracle performance, denoted as \textbf{Gap Recovered}. We find that the gains associated with this are comparable to or larger than the gains associated with Best-of-$N$ using the same evaluation strategy on the outputs of instruction-tuned generators.}
\label{tab:self_ver_bon}
\end{table*}

\begin{table*}[!t]
\centering
\footnotesize
\resizebox{0.7\textwidth}{!}{
\begin{tabular}{l c cc}
\toprule
\textbf{Evaluator} & \textbf{\# Candidate Responses} & \textbf{AIME-2024} & \textbf{AMC-2023} \\
\midrule
Greedy & 1 & 50.0 & 85.54 \\
Skywork-8B (ORM) & 64 & 60.0 & 77.10 \\
Qwen-PRM-7B & 64 & 40.0 & 73.50 \\
Skywork-27B (ORM) & 64 & 60.0 & 88.00 \\
Qwen-PRM-72B & 64 & 56.66 & 85.54 \\
DeepSeek-R1-Distill-Qwen-7B (Ours; Process + Outcome) & 8 & \textbf{66.7} & \textbf{89.2} \\
\bottomrule
\end{tabular}}
\caption{\textbf{Reasoning models can self-evaluate its response more effectively than using direct evaluators:} Comparison of direct evaluators (ORMs and PRMs) with 64 candidate responses versus our reasoning process evaluator with only 8 responses. We generate the responses with DeepSeek-R1-Distill-Qwen-7B and test different evaluators in Best-of-$N$ setting. Our approach achieves the highest accuracy on both benchmarks.}
\label{tab:self_ver_bon2}
\end{table*}

In \autoref{section:problem_solving}, we study whether using reasoning models as Best-of-$N$ evaluators improves the problem-solving capabilities of instruction-tuned models. This naturally raises the question of whether similar gains could be achieved if a reasoning model is used both as the generator and the evaluator (\textit{i.e.}, \textbf{self-evaluation}). To explore this, we conduct a preliminary experiment in which DeepSeek-R1-Distill-Qwen-7B is used both to generate responses and to evaluate them. Due to computational constraints, we only assess self-evaluation on AIME24 and AMC23 in the Best-of-8 setting.

\paragraph{The challenge of assessing long CoTs} 
A notable characteristic of current reasoning models is that their CoTs (bookended by ``\texttt{<think></think>}'' tokens) are often lengthy and include numerous reasoning steps. This presents practical challenges for evaluation, as the evaluator must be able to handle long contexts and accurately assess text that includes exploratory reasoning, backtracking, and self-correction steps. However, rather than evaluating the entire CoT trace, we could instead evaluate the reasoning \textbf{summary} that is automatically produced by the reasoning model after the CoT. This reasoning summary condenses the exploratory CoT into a more concise form resembling the CoT of instruction-tuned models: see Appendix \ref{sec:reasoning_example} for an illustrative example.

\paragraph{Experimental setting} We first generate responses to the AIME24 and AMC23 datasets with DeepSeek-R1-Distill-Qwen-7B, setting $t=0.6$ and $N=8$. We then perform Best-of-$N$ by prompting DeepSeek-R1-Distill-Qwen-7B to act as a reasoning outcome + process evaluator, following the method described in \autoref{subsec:beam}. In addition to reporting Best-of-$1$ and Best-of-$8$ performance, we also report the percentage of the performance gap between the Best-of-$1$ and oracle performances recovered by Best-of-$8$ (denoted as \textbf{Gap Recovered}). For self-evaluation, we always perform reasoning process evaluation on the output summaries, whereas we experiment with reasoning outcome evaluation on both the summaries and the entire CoT. We document our findings in Table \ref{tab:self_ver_bon}.

\paragraph{Main Results} Our results in Table \ref{tab:self_ver_bon} provide preliminary evidence that reasoning models can be used to improve their own outputs through Best-of-$N$. Specifically, the gains associated with this (Gap Recovered for DeepSeek-R1-Distill-Qwen-7B) is comparable to or larger than the gains associated with Best-of-$N$ on the outputs of the other generators (Eurus-2-SFT, Llama-3.1-70B-Instruct, and Qwen2.5-7B-Instruct) when using the same evaluator (DeepSeek-R1-Distill-Qwen-7B). 

\paragraph{Evaluating the summary is as effective as evaluating entire CoT} We also find that performing the outcome evaluation component of our reasoning process + outcome evaluation strategy on thoughts improves over outcome evaluation on summaries for AIME, whereas the opposite is true for AMC, although both strategies achieve notable gains over Best-of-1 on both datasets. We hope that our results encourage further investigation into  self-evaluation strategies for reasoning models.

\paragraph{Comparison with direct evaluators} We further evaluate the performance of state-of-the-art ORMs and PRMs under the same setting, fixing the generator to DeepSeek-R1-Distill-Qwen-7B and varying only the evaluators. While baseline evaluators operate on 64 candidate responses, our reasoning-based evaluator uses only 8 responses from the generator. The last row of \autoref{tab:self_ver_bon2} corresponds to our proposed method, which follows the ``Outcome Eval on summary + Process Eval on summary'' configuration reported in \autoref{tab:self_ver_bon}.

The results demonstrate that generating a moderate number of responses ($N=8$) with a reasoning model and evaluating them with a reasoning evaluator achieves higher accuracy than generating a much larger set of responses ($N=64$) evaluated with ORMs or PRMs. This finding highlights the complementary role of reasoning: the reasoning model not only improves problem solving through its longer chains of thought, but also enhances evaluation fidelity. Taken together, these results confirm that using reasoning models for both generation and evaluation provides a synergistic effect that surpasses conventional test-time scaling approaches.

\newpage

\subsection{Example of a reasoning model's output}\label{sec:reasoning_example}

In \autoref{fig:reasoning_example}, we provide an example of DeepSeek-R1-Distill-Qwen-7B's response to an input from AMC23. The pink box contains the chain-of-thought portion of the response, while the green box contains the summary. The CoT and the summary are separated by a ``\texttt{</think>}'' token.\\

\clearpage

\begin{figure*}[t]
\begin{tcolorbox}[colback=white, colframe=black]

\textbf{Problem.} In the state of Coinland, coins have values $6,10,$ and $15$ cents. Suppose $x$ is the value in cents of the most expensive item in Coinland that cannot be purchased using these coins with exact change. What is the sum of the digits of $x?$

\vspace{1em}
\textbf{Response}
\vspace{0.5em}

\begin{tcolorbox}[colback=pink!15, colframe=black, title=Chain-of-thoughts]
Okay, so I have this problem about Coinland where the coins are 6, 10, and 15 cents. I need to find the largest amount of money that can't be made using these coins, and then find the sum of the digits of that amount. Hmm, interesting.

\vspace{0.5em}
First, I remember something about the Frobenius number, which is the largest number that can't be expressed as a combination of given coin denominations. But I think that's mainly for two coins. Since we have three here, it might be more complicated. Maybe I should try to see if there's a pattern or a way to approach it.

\vspace{0.5em}
Let me list out some numbers and see which ones can be made with 6, 10, and 15. Maybe I can find a point beyond which all numbers can be made, and then the largest one before that would be the answer.

\vspace{0.5em}
Starting from 1 cent and going upwards:

\vspace{0.5em}
...

\vspace{0.5em}
So, starting from 30, every number can be made. Therefore, 29 is indeed the largest number that cannot be formed.

\vspace{0.5em}
So, the answer is 29, and the sum of its digits is $2 + 9 = 11$.

\vspace{0.5em}
\textbf{Final Answer}\\
The sum of the digits of the largest non-makable amount is $\boxed{11}$. \texttt{</think>}
\end{tcolorbox}

\begin{tcolorbox}[colback=green!10, colframe=black, title=Summary]
To determine the largest amount of money that cannot be made using coins of 6, 10, and 15 cents, we start by listing the numbers and checking if they can be formed using these coins.

\vspace{0.5em}
We find that the numbers 1, 2, 3, 4, 5, 7, 8, 9, 11, 13, 14, 17, 19, 23, and 29 cannot be formed. We then check numbers beyond 29 and find that all numbers from 30 onwards can be formed using combinations of 6, 10, and 15 cents.

\vspace{0.5em}
Thus, the largest number that cannot be formed is 29. The sum of the digits of 29 is $2 + 9 = 11$.
\[
\boxed{11}
\]
\end{tcolorbox}

\end{tcolorbox}
\caption{We provide an example of DeepSeek-R1-Distill-Qwen-7B's response to an input from AMC23. The pink box contains the chain-of-thought portion of the response, while the green box contains the summary. The CoT and the summary are separated by a ``\texttt{</think>}'' token.}
\label{fig:reasoning_example}
\end{figure*}

\clearpage



\section{Additional analyses (\autoref{sec:discussion}}

\subsection{When does process evaluation fail?}

\begin{figure}
    \centering
    \includegraphics[width=\linewidth]{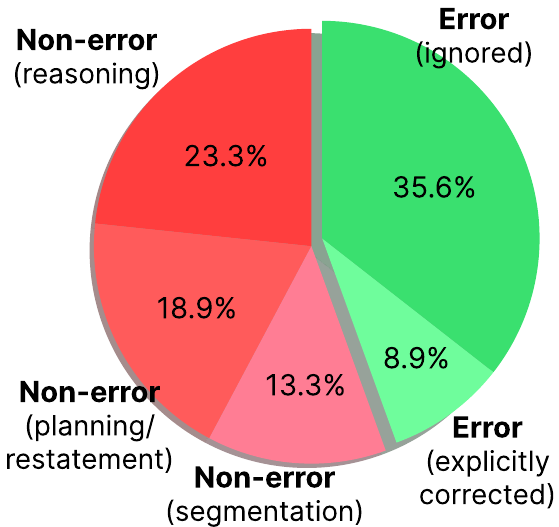}
\caption{Error pattern analysis of false positive cases from reasoning process evaluators. Please see \autoref{sec:appendix-manual} for more details.}
\label{fig:manual-analysis}
\end{figure}

\textbf{When does process evaluation fail?} We manually analyze 90 false negative cases from the 7B reasoning process evaluator on MATH-500, OlympiadBench, and GPQA. We identify two common failure modes: (1) the evaluators incorrectly flag correct steps as errors, and (2) the solutions contain actual reasoning errors despite reaching correct conclusions, a phenomenon known as \textit{unfaithful reasoning}~\citep{lyu2023faithful, wang2025examining}. Our analysis reveals that 44.4\% of flagged steps contain genuine errors (see \autoref{fig:manual-analysis}), confirming that unfaithful reasoning significantly contributes to the discrepancy between process and outcome evaluation. Further analyses with examples can be found in \autoref{tab:error_types}. 

\vspace{3mm}
\subsection{Precision-recall curve}
\label{appendix:prcurve}

\begin{figure*}[hb]
    \centering
    \includegraphics[width=1\linewidth]{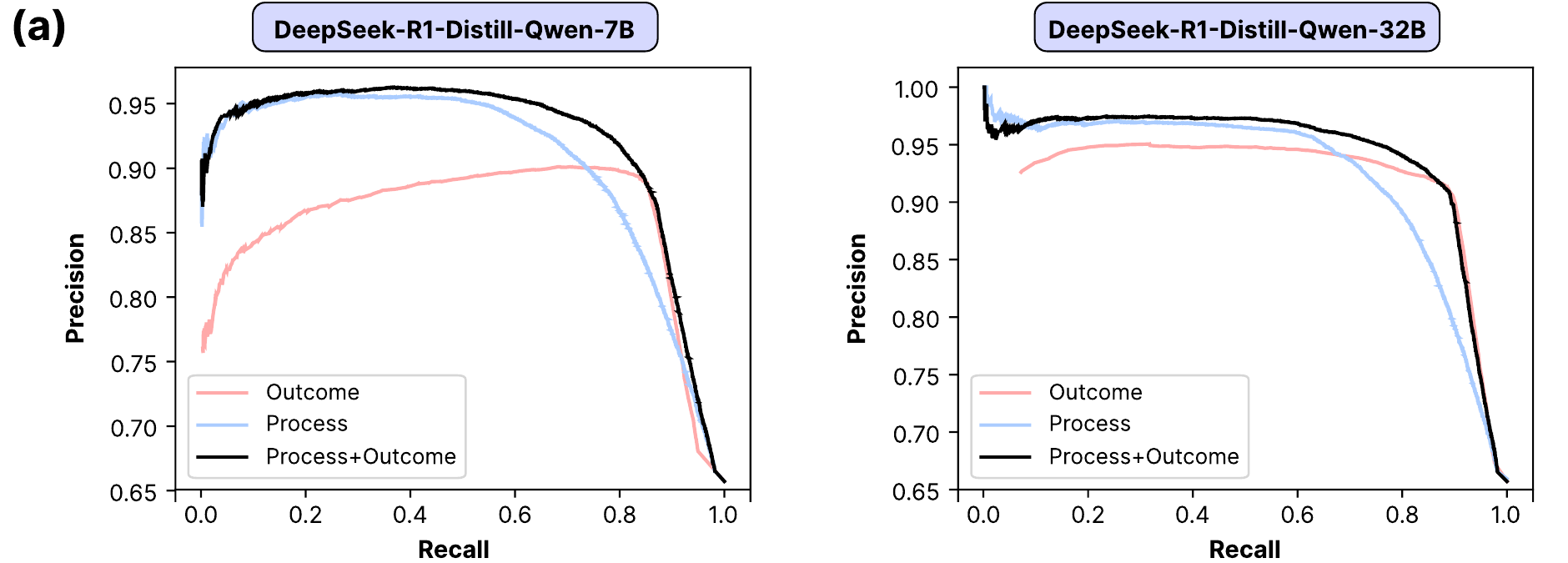}
    \caption{P-R curve of reasoning outcome evaluators and process evaluators. While reasoning process evaluators achieve higher precision in the low-recall region, reasoning outcome process evaluators achieve better performance in the high-recall region. Combining both scores is an optimal strategy that achieves better precision than outcome and process evaluation in any region.}
    \label{fig:prcurve}
\end{figure*}

\autoref{fig:analyses_figure} (Bottom) displays the confusion matrix of reasoning outcome evaluators and process evaluators using a constant classification threshold (0.5). However, one can also plot a precision-recall curve (P-R curve) by varying the threshold. In this plot, recall=0 indicates that the threshold is high ($\sim1.0$) and all correct responses are classified as negative, and recall=1 indicates that the threshold is small ($\sim0.0$) and all correct responses are classified as positive.

The resulting P-R curve is shown in \autoref{fig:prcurve}. Aligned to the findings of \autoref{sec:discussion}, the curves show that reasoning process evaluators are more \textit{conservative} classifiers than reasoning outcome evaluators. The curve indicates that if the process evaluator assigned a high score, it is more likely that the final answer is correct (high precision in low-recall region). However, outcome evaluators achieve better overall accuracy (high precision in high-recall region), indicating that process evaluators are more likely to assign relatively low scores to correct responses than outcome evaluators. Combining outcome and process scores results in similar trends to process evaluators in the low-recall region and to outcome evaluators in the high-recall region, achieving the best of both worlds.

\subsection{Qualitative analysis}
\label{sec:appendix-manual}

In this section, we briefly present the criteria for manual analysis on \textit{steps that lead to a correct answer but are predicted as incorrect by reasoning process evaluators}.

First, we randomly sample 90 responses from MATH500, OlympiadBench, and AMC generated using \texttt{Llama-3.1-70B-Instruct}, where the response's final answer is correct but the reasoning process evaluator (DeepSeek-R1-Distill-7B) flag a step-level error. The three datasets were chosen because (1) they cover a diverse range of problems including relatively easier (MATH500), medium-level (AMC), and hardest problems (OlympiadBench)\footnote{Despite that AIME is the hardest dataset with the lowest average scores, the small size (30 problems) makes it insufficient for manual error analysis.}, and (2) these three datasets demonstrate the most significant gap between reasoning outcome evaluators and reasoning process evaluators in Best-of-$N$ setting.

The authors manually analyzed the first erroneous step flagged by the reasoning process evaluators. The flagged steps are classified into \textit{errors} and \textit{non-errors}. Errors include clear logical or mathematical errors or unjustified falsifiable statements, whereas non-errors include correct reasoning steps, assumptions, and text unrelated to reasoning. The taxonomy is displayed within Table \ref{tab:error_types}.

\begin{table*}[h]
    \centering
    \scriptsize
    \begin{tabular}{|p{2.5cm}|p{11cm}|}
        \hline
        \rowcolor{gray!30} \textbf{Error Type} & \textbf{Example} \\
        \hline
        \cellcolor{green!10} \textbf{Error} \newline (ignored) \newline 35.6\% & 
        \textbf{Problem.} In the $xy$-plane, a circle of radius $4$ with center on the positive $x$-axis is tangent to the $y$-axis at the origin, and a circle with radius $10$ with center on the positive $y$-axis is tangent to the $x$-axis at the origin. What is the slope of the line passing through the two points at which these circles intersect? The final answer can be written in the form $\frac{m}{n}$, where $m$ and $n$ are relatively prime positive integers. What is $m+n$? \newline
        ... \newline
        \textbf{Step 5.} When $y = 0$, $x = 0$. This is the \textbf{origin, which is not the intersection point of the two circles} (\textit{cf. two circles intersect at the origin}), so we discard this solution. When $y = \frac{80}{29}$, $x = \frac{5y}{2} = \frac{5 \cdot 80}{2 \cdot 29} = \frac{200}{29}$. So, the coordinates of the intersection point are $(\frac{200}{29}, \frac{80}{29})$. \\ \hline
        
        \cellcolor{green!10} \textbf{Error} \newline (explicitly corrected) \newline 8.9\% & 
        \textbf{Problem.} Zach has three bags and a bunch of pencils to be placed into the bags. He is told to place the greatest number of pencils possible into each of the three bags while also keeping the number of pencils in each bag the same. What is the greatest number of pencils he could have left over? \newline
        ... \newline
        \textbf{Step 2.} Since Zach has to distribute the pencils evenly among the three bags, the total number of pencils must be a multiple of 3. However, \textbf{to have the greatest number of pencils left over, we need to consider the scenario where the total number of pencils is 1 more than a multiple of 3.} (\textit{cf. corrected by Step 5})\newline
        \textbf{Step 5.} However, we are looking for the greatest number of pencils left over. Since 3x + 2 is the largest multiple of 3 plus a remainder, the greatest number of pencils Zach could have left over is 2.
        \\ \hline
        
        \cellcolor{red!10} \textbf{Non-error} \newline (reasoning) \newline 23.3\% & 
        \textbf{Problem.} What is the value of $(3x-2)(4x+1)-(3x-2)4x+1$ when $x=4$? \newline
        ... \newline
        \textbf{Step 2.} First, substitute x = 4 into the given expression: $(3(4)-2)(4(4)+1)-(3(4)-2)4(4)+1$ \\ \hline
        
        \cellcolor{red!10} \textbf{Non-error} \newline (planning/restatement) \newline 18.9\% & 
        \textbf{Problem.} There are two increasing sequences of five consecutive integers, each of which have the property that the sum of the squares of the first three integers in the sequence equals the sum of the squares of the last two. Determine these two sequences. \newline
        ... \newline
        \textbf{Step 1.} To solve this problem, let's break it down step by step. Let the first sequence be: a, a+1, a+2, a+3, a+4 \\ \hline
        
        \cellcolor{red!10} \textbf{Non-error} \newline (segmentation) \newline 13.3\% & 
        \textbf{Problem.} Suppose that $f(a)=2 a^{2}-3 a+1$ for all real numbers $a$ and $g(b)=\log _{1/2} b$ for all $b>0$. Determine all $\theta$ with $0 \leq \theta \leq 2 \pi$ for which $f(g(\sin \theta))=0$. \newline
        ... \newline
        \textbf{Step 9.} This gives us two possible values for $u$: $u = 1/2$ and $u = 1$. \newline
        \textbf{Step 10.} Now, we substitute back in for $u$. $log_{1/2} (\sin \theta) = 1/2$ $log_{1/2} (\sin \theta) = 1$\newline (\textit{cf.} $log_{1/2} (\sin \theta) = 1/2$ \texttt{\textbackslash n} $log_{1/2} (\sin \theta) = 1$)\\ \hline
        
        \hline
    \end{tabular}
    \caption{Examples of steps flagged as an error by reasoning process evaluators.}
    \label{tab:error_types}
\end{table*}

\clearpage

\section{Analysis: How do mixing process and outcome evaluation lead to superior performance?}\label{appendix:why}

\begin{figure*}[!t]
    \centering
    \includegraphics[width=0.8\linewidth]{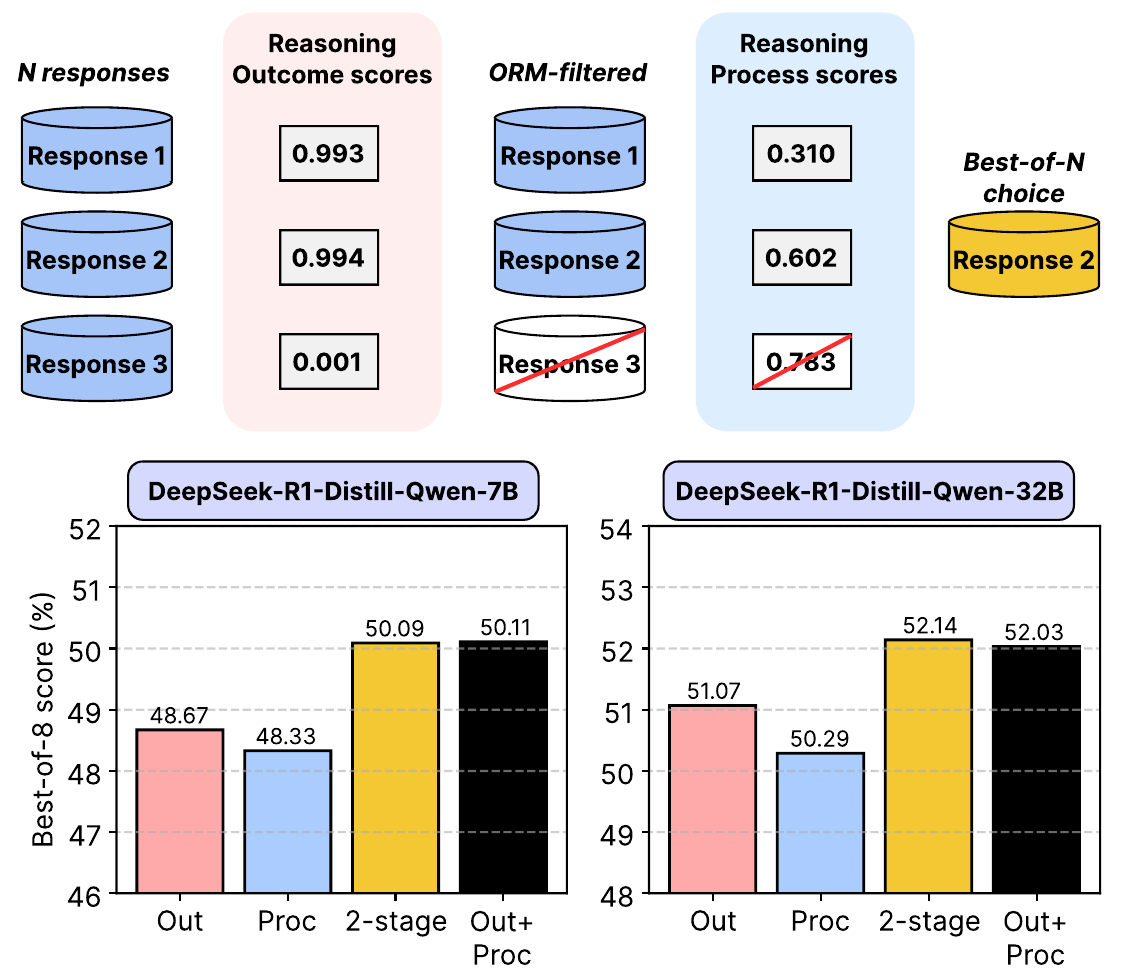}
\caption{\small{While reasoning outcome evaluators are generally better at finding the correct answer due to high recall, reasoning process evaluators can perform tie-breaking with high accuracy among outcome evaluator-filtered samples, outcome scores, and even process+outcome scores. This suggests that process evaluators can efficiently filter false positives, \textit{i.e.}, the responses that outcome evaluators classified as correct but contains process-level errors.}}
\label{fig:analyses_q2}
\end{figure*}

\begin{figure}
    \centering
    \includegraphics[width=\linewidth]{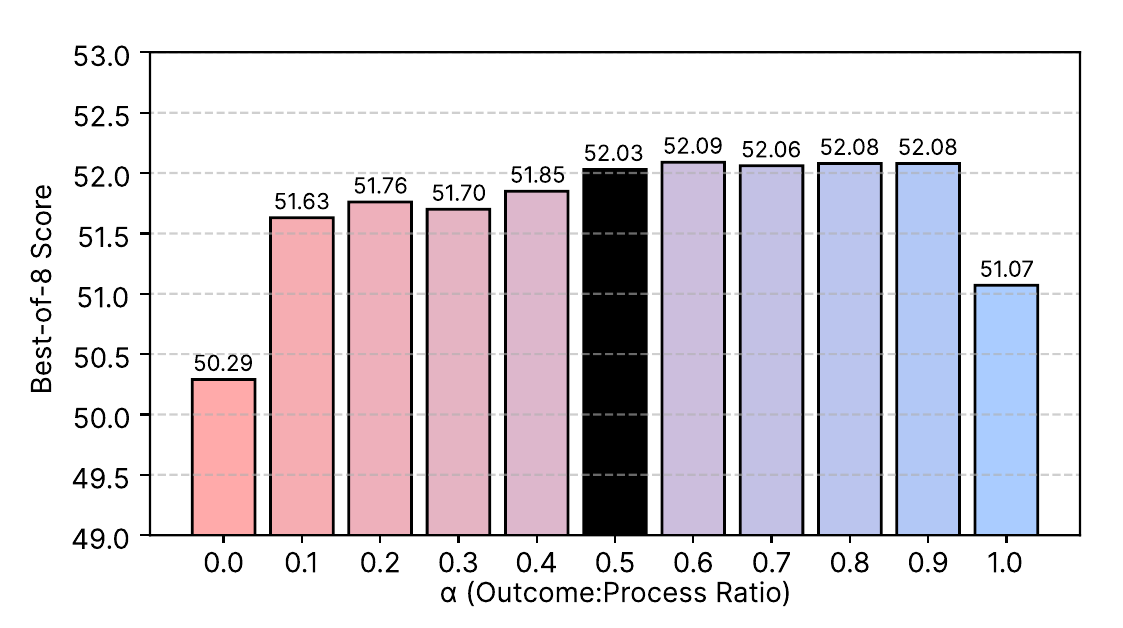}
\caption{\small{Optimal mixing rate between reasoning outcome scores and process scores is skewed towards outcome evaluator. Increasing the proportion of process scores leads to reduced scores.}}
\label{fig:mixing_rate}
\end{figure}

Our reasoning process + outcome evaluator baseline averages the scores from reasoning process evaluators and reasoning outcome evaluators and has shown to be effective in our Best-of-$N$ experiments at \autoref{section:problem_solving}. To better understand the reasons behind this, we first analyze how the results change when mixing with different ratios ($\alpha$ values). We perform a grid search of the $\alpha$ value from $0.0$ to $1.0$ with step size $0.1$ and find that the optimal $\alpha$ is skewed towards the outcome score, where weighting process score more than 0.5 causes the performance to decline. (Figure \ref{fig:mixing_rate}) \footnote{Throughout this section, the reported scores are from DeepSeek-R1-Distill-Qwen-32B.}

The optimal mixing rate highly (but not entirely) skewed towards outcome evaluators suggests that process evaluation serves as a \textbf{tie-breaker} for outcome evaluation when merged. Since reasoning outcome evaluators output tokens \texttt{0}/\texttt{1} as the correctness label, the scores (token probabilities of the label 1) are indistinguishable between responses labeled as correct or wrong. In process+outcome evaluators, process scores can be applied to break ties in responses by penalizing process errors, leading to improved Best-of-$N$ accuracy.

To prove this intuition that process evaluators can further rerank responses indistinguishable by outcome evaluators, we explore an alternative of $\alpha$-weighted average version of process+outcome evaluators, \textbf{2-stage prompting} (Figure \ref{fig:analyses_q2}). In this setting, responses are first filtered using the outcome score. Responses with outcome scores higher than 0.99 were analyzed by process evaluators, selecting the top response. Therefore, responses with \textit{low outcome scores but high process scores} cannot be chosen as the final candidate. Intuitively, the 2-stage prompting's performance is strictly bounded by outcome evaluator's recall and process evaluator's precision, whereas the soft merging of the process+outcome evaluator offers more flexibility.

The results show that the performance of the 2-stage prompting is significantly higher than that of outcome evaluator and is almost identical to that of process+outcome evaluator. As the difference between reasoning outcome scores is extremely small, using only outcome scores might not entirely reflect the quality of the responses and lead to a suboptimal Best-of-$N$ performance. However, process evaluators can further rerank responses that outcome evaluators assign indistinguishable scores as shown in 2-stage prompting, which is the key aspect of the optimal Best-of-$N$ performance of reasoning process+outcome evaluator.

One benefit of 2-stage prompting is that it reduces the inference cost by only applying the reasoning process evaluation to responses that passed the outcome evaluation. While different heuristics can be applied to optimize the compute while retaining the Best-of-$N$ performance (\textit{e.g.} perform process evaluation only if outcome evaluators classified responses with different final answers as correct), we leave this direction as future work.

\begin{figure*}[!t]
    \centering
    \includegraphics[width=1\linewidth]{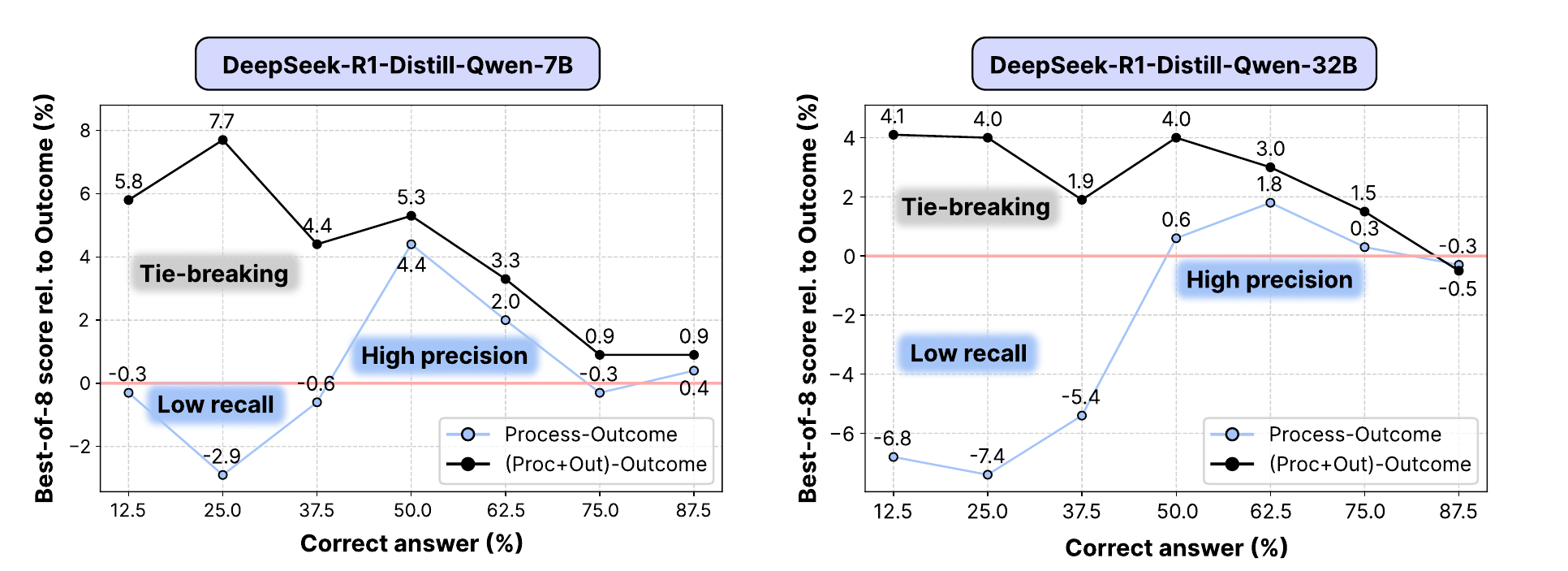}
    \caption{While reasoning process evaluators achieve low Best-of-$N$ score compared to reasoning outcome due to low recall (\autoref{sec:discussion}), Reasoning process+outcome evaluators outperform outcome evaluators by leveraging the tie-breaking ability of process evaluators. Both effects are more significant in difficult problems, where the response generator models are unlikely to find the correct answer.}
    \label{fig:difficulty_plot}
\end{figure*}

\section{Analysis: How does problem difficulty affect outcome and process evaluation?}
\label{appendix:difficulty}

Another important factor regarding Best-of-$N$ performance is the \textit{problem difficulty}, often estimated by the fraction of correct answers out of $N$ responses. The fraction value is empirically important because if there are more correct answers, there is a higher chance of selecting a response with a correct answer. However, if there are only a few correct answers, it is generally challenging to rank the correct answer at the top.

As seen in the relative performance (Figure \ref{fig:difficulty_plot}), Reasoning outcome evaluators outperform process evaluators in difficult problems, whereas process evaluators achieve higher Best-of-$N$ accuracy in relatively easier problems. This can be explained by the conclusion of \autoref{appendix:why}, that reasoning process evaluators are conservative classifiers and often assigns low score to responses with correct answers. However, if there is a sufficient amount of correct responses, the conservative nature of process evaluators prevents choosing responses with incorrect steps, increasing the expected quality of the top response.

The problem difficulty also affects the performance gap between reasoning process+outcome evaluators and reasoning outcome evaluators. Fewer correct answers increase the chance of \textit{false positives} in outcome evaluators, where they assign high (>0.99) scores to responses with incorrect answers. When using process scores together, such false positives can be effectively reranked and filtered as shown in \autoref{appendix:why}, leading to improved performance in the Best-of-$N$ setting.

\section{Prompts for Reasoning Evaluators}\label{appendix:reasoning_evaluator_prompts}

We include the prompts used to elicit reasoning models as process and outcome evaluators:

\clearpage

\fbox{
    \parbox{\textwidth}{\textbf{Reasoning process evaluator prompt:} \\~\\
    The following is a math problem and a solution (split into paragraphs, enclosed with tags and indexed from 0): \\~\\

\textbf{Problem}

\{problem\}
\\~\\
\textbf{Previous Paragraph(s)}

\{previous\_paragraphs\} 
\\~\\
\textbf{Current Paragraph}

\{current\_paragraph\}
\\~\\
\textbf{Instructions}

Your task is to decide whether the current paragraph is correct or not. If the current paragraph is correct, return the index of 1 and if not, return the index of 0.
\\~\\
Don't try to solve the problem. Your task is only to critique the current paragraph.
\\~\\
Please put your final prediction (i.e., the correctness, which must be 0 or 1) in boxed\{\{\}\}. Every output must therefore contain either \boxed{{1}} or \boxed{{0}}.
\\~\\
You should only consider the logical correctness of the current paragraph, not whether it is useful or has the potential to lead to the correct answer.
\\~\\
    }
}

\clearpage

\fbox{
    \parbox{\textwidth}{\textbf{Reasoning outcome evaluator prompt:} \\~\\
    The following is a math problem and a solution (split into paragraphs, enclosed with tags and indexed from 0): \\~\\

\textbf{Problem}

\{problem\}
\\~\\
\textbf{Response}

\{response\}
\\~\\
\textbf{Instructions}

Your task is to decide whether the solution is correct or not. If the solution is correct, return the index of 1 and if not, return the index of 0.
\\~\\
Don't try to solve the problem. Your task is only to critique the solution.
\\~\\
Please put your final answer (i.e., the index, which must be 0 or 1) in boxed\{\{\}\}. Every output must therefore contain either \boxed{{1}} or \boxed{{0}}.
    }
}

\clearpage

\section{Licenses}
\label{appendix:license}

\subsection{Datasets}

We disclose the licenses of the datasets used in this study, as indicated in their official HuggingFace repository (if applicable).


\begin{itemize}
    \item ProcessBench: Apache 2.0
    \item GSM8k: MIT
    \item MATH: MIT
    \item AIME24: CC0 1.0
    \item AMC23: License type not mentioned. Copyright © Mathematical Association of America. 
    \item MinervaMath: MIT (LM-eval-harness)
    \item OlympiadBench: Apache 2.0
    \item LeetCode: MIT
    \item GPQA: MIT
\end{itemize}

The datasets used in this study does not contain personally identifying information or offensive content.

\subsection{Models}

We disclose the licenses of the model used in this study, as indicated in their official HuggingFace repository.

\begin{itemize}
    \item Llama 3 family and their derivatives: Meta Llama 3 Community License. Copyright © Meta Platforms, Inc. All Rights Reserved.
    \item Qwen family and their derivatives: Qwen License. Copyright © Alibaba Cloud. All Rights Reserved.
    \item QwQ family: Apache 2.0
    \item DeepSeek-R1 family: MIT
    \item Skywork family: Skywork Community License
    \item Prometheus 2 family: Apache 2.0
\end{itemize}

\section{Compute Resources}\label{appendix:gpu_info}

To conduct our experiments with 7B and 32B models we used as reasoning evaluators, we used a single node of A6000 GPUs (8 GPUs) where each GPU has 48GB of memory.

\section{Potential Risks}\label{appendix:broader_implication}

One fundamental issue with language model evaluators is that they were developed to automate human evaluation, which is expensive and time-consuming. However, verifying that these evaluators function as intended, even partially, remains crucial. In our paper, we only tested benchmarks related to mathematics and coding, but this verification becomes even more important when applying reasoning evaluators to assess language model responses on socially sensitive real-world queries or in safety-related domains. We hope that all researchers and practitioners using LM evaluators will consider these points.

\section{Use of Large Language Models}

We have used LLMs for writing this paper. Specifically, we have used it to fix grammar and enhance fluency.

\end{document}